\documentclass{llncs}

\usepackage[pdftex]{graphicx}

\usepackage{times}
\usepackage[english]{babel}
\usepackage{epsfig}
\usepackage{graphicx}
\usepackage{amssymb,amsmath}
\usepackage{amsfonts}

\newcommand{\as}{``}

\newcommand{\be}{\begin{em}}
\newcommand{\ee}{\end{em}}
\newcommand{\bb}{\begin{bf}}
\newcommand{\eb}{\end{bf}}
\newcommand{\I}[1]{\relax\ifmmode\mbox{\it#1}\else{\it#1}\fi}

\newcommand{\tbs}{\hspace*{4mm}}

\newcommand{\tbl}{\hspace*{12mm}}
\newcommand{\no}{not\,}

\newcommand{\meno}{\medskip\noindent}

\newcommand{\rif}{~\ref}
\newcommand{\IF}{\mbox{:-}}
\newcommand{\op}{\mathit{Op}}

\newcommand{\THEN}{\mbox{${:}{>}$}}
\newcommand{\WH}{\mbox{${:}{<}$}}

\newcommand{\X}{X}
\newcommand{\G}{G}
\newcommand{\F}{F}

\newcommand{\N}{N}

\newcommand{\SOMETIMES}{E}

\newcommand{\ev}{\cal{Y}}
\newcommand{\ii}{{\cal{I}}}
\newcommand{\la}{{\cal{L}}}
\newcommand{\now}{{\mathit{now}}}
\newcommand{\ag}{{\mathit{Ag}}}
\newcommand{\maag}{{\cal{M}}^{\ag}}

\newcommand{\repair}{\eta}
\newcommand{\improv}{\xi}
\newcommand{\assertr}{\mathit{assert_m(\rho)}}
\newcommand{\retractr}{\mathit{retract_m(\rho)}}

\newcommand{\calsevp}{{{\cal{S}}^{{\cal{E}}vp}}}
\newcommand{\calsevpp}{{{\cal{S}}^{{\cal{E}}vp}_1}}
\newcommand{\calsf}{{{\cal{S}}^{{\cal{F}}}}}
\newcommand{\calsff}{{{\cal{S}}^{{\cal{F}}}_1}}
\newcommand{\calsj}{{{\cal{J}}^{{\cal{J}}}}}

\begin{sloppypar}

\begin{document}


\title{Self-checking Logical Agents}

\author{
\textbf{Stefania Costantini}\inst{1}}

\institute{ Dip. di Ingegneria e Scienze dell'Informazione (DISIM), Universit\`a di L'Aquila,
Coppito 67100, L'Aquila, Italy \email{stefania.costantini@di.univaq.it}
}
\maketitle

\begin{abstract}
This paper presents a comprehensive framework for run-time self-checking of logical agents,
by means of temporal axioms to be dynamically checked.
These axioms are specified by using an agent-oriented interval temporal logic defined
to this purpose. We define syntax, semantics and pragmatics
for this new logic, specifically tailored for application to agents.
In the resulting framework, we encompass and extend our past work.
\end{abstract}

\section{Introduction}

Agent systems are more and more widely used in real-world applications: therefore,
the issue of verification is becoming increasingly important, as discussed for instance in \cite{survey-logag07}
(and in the many references therein).

According to \cite{Wooldridge01},
given representations of an agent, an environment,
and a task we wish the agent to carry out in this environment,
verification tries to determine whether
the agent will carry out the task successfully.
In particular, given the specification of agent $\mathit{Ag}$,
environment $\mathit{Env}$ and property $\psi$,
the \as verification
problem'' can be defined as the decision problem
related to establishing whether $\psi$ is reached in every run of
$\mathit{Ag}$. As discussed in \cite{Wooldridge01},
the complexity of $\psi$ will affect the complexity of this problem.
As \cite{Wooldridge01} points out, $\psi$ can be either an
\emph{achievement goal}, i.e., a desirable state the the agent wants
to reach, or a \emph{maintainance goal}, related to undesirable states
that the agent wishes to avoid.
In other words, and taking time into account, two kinds of temporal properties
can be distinguished:
\emph{liveness} properties concern the progress that an agent makes and express that a
(good) state eventually will be reached;
and \emph{safety} properties express that some (bad) state will never
be entered.

Static verification of agent programs and systems (see Section\rif{comparison}),
i.e., verification performed prior to agent activation,
can be accomplished through model-checking techniques \cite{Clarke:mc}, abstract interpretation
\cite{abstr-int} or theorem proving.
This paper presents instead an approach to dynamic (run-time) verification
of agent systems.
Our motivation is that agents behavior is affected by their interaction with the external world,
so in most practical cases the actual arrival order of events
and thus the actual agent evolution is unforeseeable. Often,
the set of possible events is so large that computing all combinations
would result in a combinatorial explosion, thus making \as a priori'' (static) verification techniques
only partially applicable. Moreover, the set of in-coming events
may be only partially known in advance, at least if one admits that agents
should learn, in the sense of enlarging over time their set of behaviors.

Therefore, we believe that static verification should be integrated with dynamic
self-checking, basically aimed at detecting
violations of wished-for properties.
A crucial point of our proposal is that, in case of violation,
agents should try to restore an acceptable or desired state of affairs
by means of run-time \emph{self-repair}.
Even in case desired properties are fulfilled, by examining relevant
parameters of its own activities an agent might apply forms
of \emph{self-improvement} so as to perform better in the future.

Self-repair and improvement are seen in the direction of overcoming or at least alleviating
\as brittleness'', that can be intended as the propensity of an agent to perform poorly
or fail in the face of circumstances not explicitly considered by the agent's designer.
In opposition to brittleness, \cite{Brachman06} mentions \emph{Versatility} as the ability
of being trained to perform previously unanticipated tasks.
\cite{PerlisJLC2005,AndersonFJOPWW08} introduce the concept of \emph{Perturbation Tolerance}, where
a perturbation is any unanticipated change,
either in the world or in the system itself, that impacts an agent's performance.

To achieve Perturbation Tolerance, \cite{PerlisJLC2005,AndersonFJOPWW08} define a time-based \emph{active logic}
and a \emph{Metacognitive Loop} (MCL), that involves a system monitoring,
reasoning and meta-reasoning about and if necessary altering its own behavior.
\cite{AndersonFJOPWW08} presents a
number of domain-specific implementations of MCL,
to demonstrate that MCL is a general-purpose methodology for building flexible
(non-brittle) programs in specific domains,
and discusses the perspective of domain-independent
implementation of MCL.

We agree with \cite{PerlisJLC2005} on the fact that \as self-consciousness''
and self-modification are key aspects for building flexible and adaptable agents.
In fact, we propose a comprehensive framework for checking the agent behavior
correctness during the agent activity, aimed at self-repair and improvement
(within this framework we encompass and extend pieces
of our previous work \cite{postdalt06,CostantiniM11,costandper}).
We define an interval temporal logic devised to this aim,
and particularly tailored to the agent realm,
called A-ILTL (for Agent-oriented Interval Temporal Logic), for which we provide
a full semantics.
Based on A-ILTL formulas, we introduce two new kinds
of constraints to be added to agent programs and to be checked dynamically, that we call, respectively,
\emph{A-ILTL Rules} and \emph{A-ILTL Expressions}. These constraints are meant to be automatically attempted at
a certain (customizable) frequency, where priorities can be established among
properties to be verified.
A-ILTL rules and expressions, also according to what has happened and to what is supposed to happen or not to happen
in the future, define properties that should hold and what should be done
if they are violated or if they are fulfilled, so that an agent
can repair/improve its own behavior.

Our approach is very general, and thus could be adopted in several logic agent-oriented languages and formalisms.
In particular, one such language is DALI
\cite{jelia02,jelia04}, which is an agent-oriented extension to prolog that we have defined and developed
in previous work (cf. \cite{dali-references} for a comprehensive list of references
about DALI, while the DALI interpreter is publicly available at \cite{webDALI}). We have experimented with DALI
in a variety of applications (see, e.g., \cite{LIRA,woa07,CMTT08,Costantini-PAAMS12}),
from which we have drawn experience and inspiration for the present work.
We have added A-ILTL rules and expressions to DALI, where
we also have prototypically implemented the approach.
DALI is in itself an `active' agent-oriented logic programming language with certain features,
in particular the \as internal events'', going towards flexible self-adaptability.
In fact, internal events allow a DALI agent to react to internal conditions
whenever they occur.
By means of internal events DALI agents can take initiatives, adopt goals
and intentions, execute plans and manipulate and revise their knowledge
on their own accord, independently of the environment.
Therefore, the new features fit gracefully in this setting.
In Section\rif{example}
we will show by means of a small but complete sample
application how A-ILTL rules and expressions can be exploited in a DALI agent.

It can be useful to remark that in the proposed framework agent's state and behavior is checked
(and possibly corrected, improved or re-arranged) during agent's functioning not by means of rules as usually intended,
but by means of special constraints which are checked automatically with frequencies and
priorities customizable according to the specific requirements of the application at hand.
This helps to alleviate the
problem mentioned in \cite{soar:comparison} that in rule-based systems
\as every item which is added to memory via a rule must be maintained by
other rules \ldots'' thus unavoidably resulting, in their opinion, in brittleness
of these system.
Brittleness and inflexibility are in fact often
attributed to rule-based systems due to
their supposed over-commitment to particular courses of action, that our approach
intends to loosen.

The paper is structured as follows. In Section\rif{evolution} we provide the reader with
some notions concerning our declarative semantics
of evolving agents, and we introduce the A-ILTL temporal logic and its semantics.
In Section\rif{ailtl} we introduce
A-ILTL Rules in various forms, and provide their semantics.
We compare the proposed approach with other approaches to agent verification
in Section\rif{comparison}. In Section\rif{example} we provide
an example demonstrating how the various elements of the proposed approach
can be put at work in synergy.
Finally, in Section\rif{end} we conclude.

\section{Agent Evolution}
\label{evolution}

\subsection{Evolutionary Semantics}

In this paper we will refer to the declarative semantics
introduced in \cite{postdalt06},
aimed at declaratively
modeling changes inside an agent which are determined both by
changes in the environment, that we call \emph{external events},
and by the agent's own
self-modifications, that we call \emph{internal events}. The key idea is to understand these changes as
the result of the application of program-transformation functions
that perform changes to the agent program.
E.g., the internal event corresponding to the decision of the agent to
embrace a goal triggers a program
transformation step, resulting in a version of the program where a
corresponding plan is \as loaded'' so as to
become executable.

An agent in this framework is defined as the tuple
$\ag$ = $<P_{\ag},E,I,A>$ where $\ag$ is the agent name and $P_{\ag}$
(that we call \as agent program'') describes
the agent behavioral rules in some agent-oriented language $\la$.
$E$ is the set of the external events,
i.e, events
that the agent is capable to perceive and recognize: let $E =$
$\{E_1,\ldots,E_n\}$ for some $n$. $I$ is the
internal events set (distinguished internal conclusions): let $I =$
$\{I_1,\ldots,I_m\}$ for some $m$. $A$ is the set of actions that the agent can possibly
perform: let $A =$
$\{A_1,\ldots,A_k\}$ for some $k$. Let $\ev$ = $(E \cup I \cup A)$.
In set $\ev$, a postfix (to be omitted if irrelevant) indicates
the kind of event. I.e., $X_E$ is an external event, $X_A$ is an action and $X_I$ an internal event.

Program $P_{Ag}$ written by the programmer is
transformed into the initial agent program $P_0$ by means of an (optional) \emph{initialization
step}, that may possibly rewrite the program in an intermediate language and/or
load a \as virtual machine'' that supports language
features and/or extract control information, etc. Thus, $P_0$ can be simply a program (logical theory) or can
have additional information associated to it.

Later on, $P_0$ will evolve
according to events that happen and actions which are performed,
through corresponding program-transformation steps (each one
transforming $P_i$ into $P_{i+1}$, cf. \cite{postdalt06}), thus producing a Program
Evolution Sequence $PE = [P_0 , \ldots , P_n , \ldots ]$. The program
evolution sequence will imply a corresponding Semantic Evolution
Sequence $ME = [M_0 , \ldots , M_n, \ldots ]$ where $M_i$ is the semantic
account of $P_i$ according to the semantics of $\la$.
Notice in fact that the approach is parametric w.r.t $\la$.

The choice of a specific $\la$ will influence the
following key points: (i) when a transition from $P_i$ to $P_{i+1}$ takes place,
i.e., which are the external and internal factors that determine a
change inside the agent; (ii) which kind of transformations are performed;
(iii) which semantic approach is adopted, i.e., how $M_i$ is
obtained from $P_i$.

Let $H$ be the \emph{history} of an agent as recorded by
the agent itself, and contains events that happened and actions that have been performed
by the agent (seen as a particular kind of event), each one time-stamped so as to indicate
when they occurred. In particular, we introduce a set P of current \as valid'' past
events that describe the current state of the world \footnote{An agent can describe the state of the world only in terms
of its perceptions, where more recent remembrances define the agent's approximation
of the current state of affairs.}, and a set PNV where to store previous ones if still useful.
Thus, the history $H$ is the couple $\langle P, PNV \rangle$.
In practice, $H$ is dynamically updated with new events that
happen:
as soon as event $X$ is perceived by the agent, it is recorded in $P$ in the form ${X_P}^Y:T_i$,
$Y \in \{E,A,I\}$.
In \cite{CostantiniM11} we have defined \emph{Past Constraints}, which allow one
to define when and upon which conditions (apart from arrival of
more recent ones) past events should be moved into PNV.

\begin{definition}[Evolutionary semantics]
Let ${Ag}$ be an agent. The evolutionary semantics
$\varepsilon^{Ag}$ of ${Ag}$ is a tuple $\langle H, PE, ME\rangle$, where $H$ is the history
of ${Ag}$, and $PE$ and $ME$ are respectively its program and semantic evolution sequences.
\end{definition}

The next definition introduces the notion of instant view of $\varepsilon^{Ag}$,
at a certain stage of the evolution (which is in principle
of unlimited length).

\begin{definition}[Evolutionary semantics snapshot]
Let ${Ag}$ be an agent, with evolutionary semantics
$\varepsilon^{Ag}$ = $\langle H, PE,
ME\rangle$. The \emph{snaphot at stage i} of $\varepsilon^{Ag}_i$ is the tuple
$\langle H_i, P_i, M_i \rangle$, where $H_i$ is the history
up to the events that have determined the transition from
$P_{i-1}$ to $P_i$.
\end{definition}

In \cite{postdalt06} we have coped in detail with evolutionary semantics
of DALI language, specifying which program transformation steps are associated
with DALI language constructs. We hope however to have convinced the reader
about the fact that the approach is in principle applicable to many other
agent-oriented languages.

\subsection{Interval Temporal Logic A-ILTL}
\label{temp-log}

For defining properties that are supposed to be respected
by an evolving system, a well-established approach is that of
Temporal Logic (introduced in Computer Science by
Pnueli \cite{Pnueli77}, for a survey the reader can refer to \cite{tl3}), and in particular of Linear-time
Temporal Logic (LTL).
LTL logics are called `linear' because, in contrast to branching time logics, they
evaluate each formula with respect to a vertex-labeled infinite path (or \as state sequence'')
$s_{0}s_{1}\ldots$ where each vertex $s_{i}$
in the path corresponds to a point in time (or \as time instant'' or \as state'').
LTL enriches an underlying propositional logic language with
a set of temporal unary and
binary connectives referring to future time and past time.
In what follows, we use the following notation
for the best-known LTL operators: $\X$ stands for `next state', or next time; $\F$ stands for
`eventually', or `sometime'; $\G$ stands for `always', $\N$ stands for `never'.

LTL expressions are interpreted in a discrete, linear
model of time. Formally, this structure is represented by ${\cal{M}} = \langle \mathbb{N}, \ii \rangle$
where, given countable set $\Sigma$ of atomic propositions, interpretation function
$\ii\,:\,\mathbb{N}$\,$\mapsto$\,${{2^{\Sigma}}}$ maps each natural number $i$ (representing state $s_i$) to a
subset of $\Sigma$. Given set $\cal{F}$ of formulas
built out of classical connectives and of LTL operators,
the semantics of a temporal formula is provided by the
satisfaction relation $\models\,:\,{\cal{M}}\,\times\,\mathbb{N}\,\times\,\cal{F}\,\rightarrow\,\{\mathit{true},\mathit{false}\}$.
For $\varphi \in \cal{F}$ and $i \in \mathbb{N}$ we write ${\cal{M}},i\,\models\,\varphi$ if,
in the satisfaction relation, $\varphi$ is true w.r.t. ${\cal{M}},i$.
We can also say (leaving ${\cal{M}}$ implicit) that $\varphi$ \emph{holds} at $i$, or equivalently in state $s_i$,
or that state $s_i$ satisfies $\varphi$.
For atomic proposition $p \in \Sigma$, we have ${\cal{M}},i$\,$\models$\,$p$ iff $p \in \ii(i)$.
The semantics of $\models$ for classical connectives is as expected, and the semantics for LTL operators is as reported in \cite{tl3}.
A structure ${\cal{M}} = \langle \mathbb{N}, \ii \rangle$ is a model of $\varphi$ if
${\cal{M}},i\,\models\,\varphi$ for some $i \in \mathbb{N}$.
Similarly to classical logic, an LTL formula $\varphi$ can be
satisfiable, unsatisfiable or valid and one can define the notions of
entailment and equivalence between two LTL formulas.

In prior work (see e.g., \cite{CDP09}) we informally introduced an extension to temporal
logic based on \emph{intervals}, called A-ILTL for `Agent-Oriented Interval LTL', that we
report, formalize and extend here.
Via A-ILTL operators the time point or the time interval in which a given temporal formula is supposed
to hold are explicitly stated. E.g., $\G_{m,n}$ ({\em always in time interval})
states that formula $\varphi$ should become true at most at state $s_m$ and then hold at least until state $s_n$.
Intervals can have an upper bound
or can be unlimited, in the sense that only the lower bound is provided.

The introduction of A-ILTL is in our opinion useful despite the fact that, since the seminal work of \cite{Koymans90},
several \as metric temporal logics'' (MTL) have been defined
(cf., e.g., \cite{AlurH91,HirshfeldR04} and the references therein).
These logics are able to express \as metric'', or quantitative
time constraints. This is important and necessary as
in many systems and applications there are properties which are not required
to hold forever or somewhere in time, but in specific time instants or intervals:
in fact, a system can be required to performed a certain task \emph{at} or \emph{by} a certain time,
or for a certain duration.
MTL logics however have been mainly devised for applications involving real-time and hybrid systems,
with an underlying 'dense' (continuous) model of time, based on real numbers.
Consequently, as pointed out in \cite{HirshfeldR04}, general results about
expressiveness, decidability and complexity are lacking as
these properties turn out to be sensitive to slight differences in
the semantics or in the choice of operators.
In contrast, A-ILTL is defined
as a simple extension of LTL, then still relying upon an underlying discrete linear model of time.
We impose strong limitations upon nesting of operators, so as to avoid having
to explicitly cope with the representation of time intervals and their interactions.
However, as we will see in the rest of the paper this simple formulation is
sufficient for expressing and checking a number of interesting properties of agent systems.

Formal syntax and semantics of \as core''
A-ILTL operators (also called below \as Interval Operators'') are defined as follows.

\begin{definition}
Set $\cal{F}$ of A-ILTL formulas is
built out out of classical connectives and of LTL operators and of the following A-ILTL operators,
where $m,n$ are positive integer numbers (with $m \leq n$), and $\varphi$ and
$\psi$ are LTL formulas (i.e., nesting of A-ILTL operators is not allowed).
\begin{itemize}
\item[]
\begin{itemize}
\item[$C(i)$] ({\em current state}). $C(i)$ is true if $s_i$ is the current state. I.e., ${\cal{M}},i\,\models\, C(i')$ iff $i = i'$.
From this operator we obtain the shorthand expression $\now$,
where $\now = t : C(t)$.

\item[$p_i$] ({\em $p$ at $i$)}. Proposition $p$ holds at time $i$. This notation transposes a propositional letter into a \as timed'' version. I.e.,  ${\cal{M}},i\,\models\, p(i)$ if ${\cal{M}},i\,\models\, p$. The short form $p_{\now}$ is a shorthand, where
 ${\cal{M}},i\,\models\, p_{\now}$ if
 ${\cal{M}},i\,\models\, p_{i}$ $\wedge$  $C(i)$.

 \item[$p_{\langle i\rangle}$] ({\em $p$ since $i$)}. Proposition $p$ holds since time $i$. This notation transposes a propositional letter into a \as timed'' version, but considers a peculiar feature of agents, where due to the interaction with the environment the agent modifies its knowledge base. There is no need to change the overall semantic framework, as function $\ii$ may account for such changes. I.e.,  ${\cal{M}},i\,\models\, p{\langle i\rangle}$ if
 ${\cal{M}},i\,\models\, p$ and $\forall i' < i$, ${\cal{M}},i'\,\not\models\, p$. The short form $p_{\langle\now\rangle}$ is a shorthand, where
 ${\cal{M}},i\,\models\, p_{\langle\now\rangle}$ if
 ${\cal{M}},i\,\models\, p_{\langle i\rangle}$ $\wedge$  $C(i)$.

\item[$\X_m$] ({\em future m-state}). $\X_m\varphi$ holds if $\varphi$ is true in the $(m+1)$-th state after the current state.
I.e., ${\cal{M}},i\,\models\,\X_m\varphi$ if ${\cal{M}},i'\,\models\,\varphi$, $i' = i+m$. This operator is relative to current state
$C(i)$, in fact it can hold or not hold depending on state $i$ that is considered. Therefore a more suitable form is $\X_m(i)$,
where the reference state is explicitly stated.

\item[$\F_m$] ({\em bounded eventually (or \as finally'')}). $\F_m\varphi$ holds if $\varphi$ is true somewhere on the path from the current state to the $(m)$-th state after the current one. I.e., ${\cal{M}},i\,\models\,\F_m\varphi$ if there exists $j$ such that $j \geq i$ and $j \leq i+m$
    and ${\cal{M}},j\,\models\,\varphi$. This operator is relative to current state
$C(i)$, in fact it can hold or not hold depending on state $i$ that is considered. Therefore a more suitable form is $\F_m(i)$,
where the reference state is explicitly stated.

\item[$\F_{m,n}$] ({\em eventually (or \as finally'') in time interval}). $\F_{m,n}\varphi$ states that $\varphi$ has to hold somewhere on the path from state $s_m$ to state $s_n$. I.e., ${\cal{M}},i\,\models\,\F_{m,n}\varphi$ if there exists $j$ such that $j \geq m $ and $j \leq n$ and
    ${\cal{M}},j\,\models\,\varphi$.

\item[$\G_{m}$] ({\em bounded always}). $\G_{m}\varphi$ states that $\varphi$ should become true at most at state $s_m$. It is different from LTL $\G$ in that the state where the property should start to hold is explicitly indicated. I.e., ${\cal{M}},i\,\models\,\G_{m}\varphi$ if for all $j$ such that $j \geq m $
    ${\cal{M}},j\,\models\,\varphi$.

\item[$\G_{\langle m\rangle}$] ({\em bounded strong always}). $\G_{m,n}\varphi$ states that $\varphi$ should become true just at state $s_m$, while it was not true at previous states. I.e., ${\cal{M}},i\,\models\,\G_{\langle m\rangle}\varphi$ if for all $j$ such that $j \geq m $
    ${\cal{M}},j\,\models\,\varphi$ and $\forall i' < m$, ${\cal{M}},i'\,\not\models\,\varphi$.

\item[$\G_{m,n}$] ({\em always in time interval}). $\G_{m,n}\varphi$ states that $\varphi$ should become true at most at state $s_m$ and then hold at least until state $s_n$. I.e., ${\cal{M}},i\,\models\,\G_{m,n}\varphi$ if for all $j$ such that $j \geq m $ and $j \leq n$
    ${\cal{M}},j\,\models\,\varphi$.

\item[$\G_{\langle m,n \rangle}$] ({\em strong always in time interval}).  $\G_{\langle m,n \rangle}\varphi$ states that $\varphi$ should become true just in $s_m$ and then hold until state $s_n$, and not in $s_{n+1}$.
    I.e., ${\cal{M}},i\,\models\,\G_{\langle m,n\rangle}\varphi$ if for all $j$ such that $j \geq m $ and $j \leq n$ ${\cal{M}},j\,\models\,\varphi$, and
    $\forall j' < m$, ${\cal{M}},j'\,\not\models\,\varphi$, and ${\cal{M}},j''\,\not\models\,\varphi$, $j'' = n+1$.

\item[$\N^b_{m}$] ({\em never before}). $\N^b_{m}\varphi$ states that $\varphi$ should not be true in any state prior than $s_m$,
i.e., ${\cal{M}},i\,\models\,\N^b_{m}\varphi$ if there not exists $j$ such that $j < m $ and
    ${\cal{M}},j\,\models\,\varphi$.

\item[$\N^a_{m}$] ({\em never after}). $\N^a_{m}\varphi$ states that $\varphi$ should not be true in any state after $s_m$,
i.e., ${\cal{M}},i\,\models\,\N^a_{m}\varphi$ if there not exists $j$ such that $j > m $ and
    ${\cal{M}},j\,\models\,\varphi$.

\item[$\N_{m,n}$] ({\em never in time interval}). $\N_{m,n}\varphi$ states that $\varphi$ should not be true in any state between $s_m$ and $s_n$,
i.e., ${\cal{M}},i\,\models\,\N_{m,n}\varphi$ if there not exists $j$ such that $j \geq m $ and $j \leq n$ and
    ${\cal{M}},j\,\models\,\varphi$.

\item[$\SOMETIMES_{m,f}$] ({\em bounded sometimes}). $\SOMETIMES_{m,n}\varphi$ states that $\varphi$ has to be true one or more times starting  from state $s_m$, with frequency $f$. I.e., ${\cal{M}},i\,\models\,\SOMETIMES_{m,f}\varphi$ if
    ${\cal{M}},m\,\models\,\varphi$ and ${\cal{M}},i\,\models\,\SOMETIMES_{m',f}\varphi$, $m' = m+f$.

\item[$\SOMETIMES_{m,n,f}$] ({\em sometimes in time interval}). $\SOMETIMES_{m,n}\varphi$ states that $\varphi$ has to be true one or more times between $s_m$ and $s_n$, with frequency $f$. I.e., ${\cal{M}},i\,\models\,\SOMETIMES_{m,n,f}\varphi$ if
    ${\cal{M}},i\,\models\,\varphi$ whevener $m+f \geq n$, or otherwise if ${\cal{M}},m\,\models\,\varphi$
    and ${\cal{M}},i\,\models\,\SOMETIMES_{m',n}\varphi$ with $m' = m+f$ .

\end{itemize}
\end{itemize}
\end{definition}

Other A-ILTL operators (also referring to the past) can be defined
(see \cite{CDP09}, where a preliminary version of A-ILTL, called I-METATEM, was presented) but we do not report them
here for the sake of brevity.
There is no need to change the notion of model reported above for LTL: in fact, drawing inspiration
from the LTL treatment of `next state' and `sometime', an A-ILTL formula holds in a state if it can be checked to hold
in the states included in the interval the formula refers to. In this sense, it is the whole state sequence
that implies truth or falsity of an A-ILTL formula. However,
it is easy to see that for most A-ILTL formulas $\op\varphi$ there is a \emph{crucial state}
where it is definitely possible to assess whether the formula holds or not in given state sequence,
by observing the sequence up to that point and ignoring the rest.
The crucial state is for instance $i+m$ for $\X_m(i)$, some $j$ with $j \geq i$ and $j \leq m$ for $\F_m(i)$,
etc. It corresponds to the
upper bound of the \emph{interval of interest} of operator $\op$, which is the
interval $[v,w]$ of the first and last states where the inner formula $\varphi$ must be checked,
according to the semantic definition of the operator. Sometimes $w$ can be $\infty$.
If $w = \infty$ then there is no crucial state, as it is the case for instance for $\N^a_m$.

In the above formulation, for simplicity we do not allow A-ILTL operators to be nested.
An extension in this sense is possible, but one should then consider how temporal intervals
interact (cf., e.g., \cite{itl2004} for a survey of many existing
interval temporal logics), and this requires relevant modifications to the semantic approach
that should explicitly cope with time intervals instead of time instants.
Instead, A-ILTL operators can occur within LTL ones. We have purposedly defined a
very simple interval logic that, as we will see below, is sufficient to our aims
with no computational extra-burden.
In fact, it is easy to get convinced that the addition of interval operators leaves the
complexity of the resulting logic unchanged with respect to plain LTL.

We can employ A-ILTL formulas in \emph{response formulas} (also called \emph{response rules}) in the sense of \cite{Manna2010}.
They are of the form $p \Rightarrow q$, meaning that any state
which satisfies $p$ must be followed by a later state which satisfies $q$
(see \cite{Manna2010} for formal properties of $\Rightarrow$).
Assuming for instance a suitable encoding of time and date with
corresponding operators for adding, say, minutes hours or days to a given date, one
can write by means of A-ILTL formulas response rules such as the following, stating that
upon an order received at date $d$, the corresponding product should be
delivered within $k$ days.
\[\mathit{received\_order}_{\langle d\rangle} \Rightarrow \F_{d{+}k_{\mathit{days}}}\mathit{deliver\_product}\]

To express that orders must \emph{always} be processed this way, we can exploit the
corresponding LTL operator:

\[G(\mathit{received\_order}_{\langle d\rangle} \Rightarrow \F_{d+k_{\mathit{days}}}\mathit{deliver\_product})\]

The A-ILTL formula below states that one (maybe an e-mail system), if realizing to be expecting new mail,
should from now on check the mailbox every 5 minutes ($5_m$).
\[G_{\now}\mathit{expect\_mail} \Rightarrow \SOMETIMES_{\now,5_m} \mathit{check\_mail}\]

A-ILTL is particularly well-suited for the agent setting, where temporal aspects
matter also from the point of view of when, since when and until when agent properties should hold.

Similar rules, but without intervals, can be expressed in the METATEM logic
\cite{barringer89metatem,Fisher05,metatem1995},
whose language is based on classical propositional logic enriched by temporal connectives
and on the direct execution of temporal logic statements: in the METATEM approach,
the $\Rightarrow$ in a response rule such as $p \Rightarrow q$ is interpreted in an imperative fashion as 'Then Do'.
I.e., a response rule is interpreted (similarly to a conditional
in traditional imperative programming languages) as 'If $p$ Then Do $q$'. In METATEM, the antecedent $p$
encodes solely (without loss of generality) properties referring to the past,
where the consequent $q$ encodes what the agent should do whenever $p$ holds.
Intervals and frequency however add significant expressive power (in the pragmatic sense,
not in reference to complexity) in practical agent settings. In fact,
these features are important and sometimes crucial in many knowledge representation
applications, including deontic defeasible logics (see, e.g., the temporal logic of violations presented in \cite{GovernatoriR11}).

\subsection{Interval Temporal Logic and Evolutionary Semantics}
\label{a-iltl-evol}

In this section, we refine our Interval Temporal Logic so as to operate on a sequence
of states that corresponds to the Evolutionary Semantics defined before.
In fact, states in our case are not simply intended as time instants.
Rather, they correspond to stages of the agent evolution, marked with the
time when each stage has been reached. Time in this setting is considered
to be local to the agent, where with some sort of \as internal clock'' is able
to time-stamp events and state changes. We borrow from \cite{Manna1991} the following
definition of \emph{timed state sequence}, that we tailor to our setting.

\begin{definition}
Let $\sigma$ be a (finite or infinite) sequence of states, where the ith state $e_i$, $e_i \geq 0$,
is the \emph{semantic snaphots at stage i} $\varepsilon^{Ag}_i$
of given agent $\ag$. Let $T$ be a corresponding sequence of time instants $t_i$, $t_i \geq 0$.
A \emph{timed state sequence for agent $\ag$} is the couple $\rho_{\ag} = (\sigma, T)$. Let $\rho_i$ be the i-th state, $i \geq 0$,
where $\rho_i$ $=$ $\langle e_i, t_i\rangle$ $=$ $\langle \varepsilon^{\ag}_i, t_i\rangle$.
\end{definition}

We in particular consider timed state sequences which are \emph{monotonic}, i.e., if $e_{i+1} \neq e_i$ then
$t_{i+1} > t_i$. In our setting, it will always be the case that $e_{i+1} \neq e_i$ as there is no point in semantically
considering a static situation: as mentioned, a transition from $e_i$ to $e_{i+1}$ will in fact occur when something
happens, externally or internally, that affects the agent.

Then, in the above definition of A-ILTL operators, it would be immediate to
let $s_i = \rho_i$. This requires however a refinement: in fact,
in our setting, A-ILTL operators are intended to be used within agent programs to define
properties to be verified during agent evolution. In this kind of use, when writing $\op_m$ or $\op_{m,n}$
parameters $m$ and $n$ are supposed to define the interval of time in which the
property is expected to hold in the program definition: then, $m$ and $n$ will not necessarily
coincide with the time instants of the above-defined timed state sequence. To fill this gap,
we introduce the following approximation.

\begin{definition}
Given positive integer number $t$, we introduce the following terminology. We indicate with
$\check{\rho}_{t}$ state $\rho = \langle \varepsilon^{Ag}, \check{t}\rangle$
where $\check{t} \geq t$ and $\forall \rho' = \langle {\varepsilon^{Ag}}', t'\rangle$ with $t' <  \check{t}$ we have $t' < t$.
We indicate with $\hat{\rho}_{t}$ state $\rho = \langle \varepsilon^{Ag}, \hat{t}\rangle$
where $\hat{t} \geq t$ and $\forall \rho'' = \langle {\varepsilon^{Ag}}'', t''\rangle$ with $t'' > \hat{t}$ we have $t'' > t$.
\end{definition}

That is, if $t$ represents a time instant not exactly coincident with an element of the
time sequence $T$, $\check{\rho}_{t}$ is the state whose time component better approximates $t$
\as by defect'', i.e., the upper bound of all
states whose time component is smaller than $t$. Symmetrically, $\hat{\rho}_{t}$ is the state whose time component better approximates $t$
\as by excess'', i.e., the lower bound of
all states whose time component is greater than $t$. Notice that, given A-ILTL expression $\op_{m,n}\,\phi$
state $\check{\rho}_{t}$ is the first state where $\phi$ becomes `observable' in the agent's semantics, i.e.,
the first state where $\phi$ is actually
required to hold, and $\hat{\rho}_{t}$ is the last such state.
Therefore in the following, by $\op_{m}$ (resp. $\op_{\langle m\rangle}$) we will implicitly
mean $\op_{\check{\rho}_{m}}$ and by $\op_{m,n}$ (resp. $\op_{\langle m,n\rangle}$) we will implicitly
mean $\op_{\check{\rho}_{m},\hat{\rho}_{n}}$.

We need to adapt the interpretation function $\ii$ to our setting. In fact, we intend to employ A-ILTL within
agent-oriented languages. In particular, we restrict ourselves to logic-based languages
for which an evolutionary semantics and a notion of logical consequence can be defined.
Thus, given agent-oriented language $\la$ at hand,
the set $\Sigma$ of propositional letters used to define an A-ILTL semantic framework
will coincide with all
ground\footnote{An expression is ground if it contains no variables} expressions of $\la$.
Each expression of $\la$ has a (possibly infinite) number of ground versions, obtained by replacing variables with constants
(from the alphabet of $\la$) in every possible way.
A given agent program can be taken as standing for its (possibly infinite) ground version.
This is customarily done in many approaches, such as for instance Answer Set Programming (see e.g., \cite{Gelfond07} and
the references therein). Notice that we have to distinguish between logical consequence in $\la$,
that we indicate as $\models_{\la}$, from logical consequence in A-ILTL, indicated above simply as $\models$.
However, the correspondence between the two notions can be quite simply stated by specifying that
in each state $s_i$ the propositional letters implied by the interpretation function $\ii$ correspond to
the logical consequences of agent program $P_i$:

\begin{definition}
Let $\la$ be a logic language. Let $\mathit{Expr}_{\la}$ be the set of ground expressions that can be built from
the alphabet of $\la$. Let $\rho_{\ag}$ be a timed state sequence for agent $\ag$, and let
$\rho_i = \langle \varepsilon^{\ag}_i, t_i\rangle$ be the ith state, with $\varepsilon^{\ag}_i = \langle H_i, P_i, M_i \rangle$.
An A-ILTL formula $\tau$ is defined over sequence $\rho_{\ag}$ if in its interpretation
structure ${\cal{M}} = \langle \mathbb{N}, \ii \rangle$, index $i \in \mathbb{N}$ refers to $\rho_{i}$, which means that
$\Sigma = \mathit{Expr}_{\la}$ and
$\ii\,:\,\mathbb{N}$\,$\mapsto$\,${{2^{\Sigma}}}$ is defined such that, given $p \in \Sigma$, $p \in \ii(i)$ iff
$P_i \models_{\la} p$. Such an interpretation structure will be indicated with $\maag$.
We will thus say that $\tau$ holds/does not hold w.r.t. $\rho_{\ag}$.
\end{definition}

In practice, run-time verification of A-ILTL properties may not
occur at every state (of the given interval). Rather, sometimes properties
need to be verified with a certain frequency, that can even be
different for different properties.
Then, we have introduced a further
extension that consists in defining subsequences of the sequence of all states:
if $Op$ is any of the operators introduced in A-ILTL and $k > 1$,
$Op^k$ is a semantic variation
of $Op$ where the sequence of states $\rho_{\ag}$ of given agent
is replaced by the subsequence
$s_0,s_{k_1},s_{k_2},\ldots$ where for each $k_r, r \geq 1$, $k_r\ mod\ k\ = 0$, i.e., $k_r = g \times k$ for some $g\geq 1$.

A-ILTL formulas to be associated to given agent can be defined within the
agent program, though they constitute an additional but separate layer.
In fact, their semantics is defined as seen above on the agent evolution $\rho_{\ag}$.
In the next sections we will review and extend previous work on such rules, and we will
provide their semantics in terms of the above-defined framework.
Rules properly belonging to the agent program $P_i$ will be called \emph{object rules},
and the set of object rules composing $P_i$ can be called \emph{object layer}. In this sense,
$\la$ can be called \emph{object language}.
The set of A-ILTL formulas associated to given agent, that represent properties
which agent evolution should hopefully fulfil, can be called \emph{check layer} and will be
composed of formulas $\{\tau_1,\ldots,\tau_l\}$. Agent evolution can be considered to be \as satisfactory''
if it obeys all these properties.

\begin{definition}
Given agent $\ag$ and given a
set of A-ILTL expressions ${\cal{A}} = \{\tau_1,\ldots,\tau_l\}$, timed state sequence $\rho_{\ag}$
is \emph{coherent} w.r.t. ${\cal{A}}$ if A-ILTL formula $G\zeta$ with $\zeta = \tau_1 \wedge \ldots \wedge \tau_n$ holds.
\end{definition}

Notice that the expression $G\zeta$ is an \emph{invariance property} in the sense of \cite{Manna1984}.
In fact, coherence requires this property to hold for the whole agent's \as life''.
In the formulation $G_{m,n}\zeta$ that A-ILTL allows for, one can express \emph{temporally limited
coherence}, concerning for instance \as critical'' parts of an agent's operation. Or also,
one might express forms of \emph{partial} coherence concerning only some properties.

An \as ideal'' agent will have a coherent evolution, whatever its interactions with the environment can be,
i.e., whatever sequence of events arrives to the agent from the external \as world''.
However, in practical situations such a favorable case will seldom be the case, unless static verification has
been able to ensure total correctness of agent's behavior. Instead, violations will occasionally occur,
and actions should be undertaken so as to attempt to regain coherence for the future.
Also, some properties will not have to be checked anyway, but only upon occurrence of certain situations.
In the following sections, we will introduce two kinds of A-ILTL \emph{rules}
(that we also call A-ILTL \emph{expressions}), we will explain their
usefulness and provide their semantics.

A-ILTL rules may include asserting and retracting object rules or sets of object rules (\as modules''). Thus,
below we provide a semantic account of such operations, that is easily defined w.r.t. the Evolutionary Semantics. The need of modifying
the agent's knowledge base has been widely discussed with respect to EVOLP \cite{jelia02:evolp,epia03:evolp}.
It has also been discussed in \cite{CDP2011}, about agents learning by \as being told'' from other
trusted agents, by exchanging sets of rules. In the present setting, we will assume to assert/retract
ground (sets of) rules. However, in general this assumption can be
relaxed as one can resort to \emph{reified} form of rules, where variables are provisionally represented by
constants (cf. \cite{bcdl:jlc00} and the references therein). In this setting, we consider
$\assertr$ and $\retractr$
as special A-ILTL operators, with the following semantics. In particular, a rule asserted at state $s_i$ will be entailed
by next state. Symmetrically, after
retract a rule will no longer be entailed, with the provision that only existing rules can be retracted.
\begin{definition}
Let $\rho$ be a rule expressed in language $\la$ and $\rho_{\ag}$ be a timed state sequence for given agent. The set of A-ILTL operators is enriched by the following.
\begin{itemize}
\item[]
\begin{itemize}
\item[]
\begin{itemize}
\item[$\assertr$] ({\em assert rule}). $\assertr$ holds if $\rho$ belongs to the agent program in the next state.
I.e., ${\cal{M}},i\,\models\,\assertr$ if ${\cal{M}},m'\,\models\,\rho$, $m' = m+1$.
\item[$\retractr$] ({\em retract rule}). $\retractr$ holds if $\rho$ belongs to the agent program in current state, and will no longer
belong to it in the next state.
I.e., ${\cal{M}},i\,\models\,\retractr$ if ${\cal{M}},m\,\models\,\rho$ and ${\cal{M}},m'\,\not\models\,\rho$, $m' = m+1$.
\end{itemize}
\end{itemize}
\end{itemize}
\end{definition}

\section{A-ILTL rules and meta-rules in Agent Programs}
\label{ailtl}

There can be different ways of exploiting A-ILTL and in general temporal logic in agent-oriented languages.
The METATEM programming language \cite{barringer89metatem,Fisher05,metatem1995}, for instance, is directly based
upon the METATEM logic: temporal operators are interpreted as
modalities, and semantics is provided accordingly. This semantics is the `core' of
an executable `imperative' language. In fact, in the authors view rules such as response rules
are `imperative' in the sense that they imply performing actions.
We embrace a different position, on the one hand because of the complexity
(it is well-known that model-checking for LTL is PSPACE-complete, see \cite{LTL-complexity1985})
but on the other hand because, as outlined in previous section, we believe that
A-ILTL expressions might constitute a check layer to be added to agent programs,
whatever the specific formalism in which agents are expressed.

How should A-ILTL expressions be checked? As mentioned, the fact that all expressions
associated to an agent program are valid is an invariance property that should hold all along
(or at least in specific, \as critical'' intervals). However, agents evolution
is discrete, so A-ILTL expressions can at most be checked at each stage
of the evolutionary semantics. Moreover, both the specific expression and the
particular application may require checks to be performed at a certain frequency.
As seen before, we explicitly associated a frequency with the operator $\SOMETIMES$:
in fact, this frequency defines it very nature. However, for the other operators we preferred
to just define timed sub-sequences, so as
to defer the specification of frequency to the particular implementation setting rather
than introducing it into the logic.
For instance, the expression (where, following a prolog-like syntax,
predicates are indicated with lower-case initial letter and variables in upper-case):
\[G\ within\_range({Temperature})\]

\noindent
should be clearly checked much more frequently if supervising a critical appliance
than if supervising domestic heating.
Thus, following the famous statement of \cite{kowalgs1979} that \as Algorithm = Logic + Control'',
we assume to associate to each agent program specific \emph{control information}
including the frequency for checking A-ILTL operators. Specifically, in the formulation below
we associate the frequency to an (optional) additional parameter of each operator.

The representation of A-ILTL operators
within a logic agent-oriented programming language can be, e.g., the one
illustrated in Table \ref{fig:tableOP}, that we have adopted in DALI, where $m$ and $n$ denote the time interval
and $k$ is the frequency. We will call this syntax \emph{practical} or also \emph{pragmatic} syntax.
When not needed, the argument corresponding to frequency can be omitted. A plain LTL operator $OP$ can be expressed by omitting all arguments.
The operator $\mathit{OP}$
on the right column of each line is called \emph{pragmatic} A-ILTL operator and is said to \emph{correspond} to A-ILTL operator $\op$
on the left column.
\begin{table}
\centering
\begin{tabular}{ | c | p{0.3cm} p{5cm} |}
\hline {\bf A-ILTL Op$^k$} & & {\bf OP(m,n;k)} \\
\hline $\now$ & & $\mathit{NOW(t)}$ \\
\hline ${X_m}^k$ & & $\mathit{NEXT(m;k)}$ \\
\hline $F_m^k$ & & $\mathit{EVENTUALLY(m;k)}$\\
\hline ${F_{m,n}}^k$ & & $\mathit{EVENTUALLY(m,n;k)}$ \\
\hline ${G_{m}}^k$ & & $\mathit{ALWAYS(m;k)}$ \\
\hline ${G_{\langle m\rangle}}^k$ & & $\mathit{ALWAYS\_S(m;k)}$ \\
\hline ${G_{m,n}}^k$ & & $\mathit{ALWAYS(m,n;k)}$ \\
\hline ${G_{\langle m,n \rangle}}^k$ & & $\mathit{ALWAYS\_S(m,n;k)}$ \\
\hline ${N^b_m}^k$ & & $\mathit{NEVER\_B(m;k)}$ \\
\hline ${N^a_m}^k$ & & $\mathit{NEVER\_A(m;k)}$ \\
\hline ${N_{m,n}}^k$ & & $\mathit{NEVER(m,n;k)}$\\
\hline ${E_{m,k}}$ & & $\mathit{SOMETIMES(m;k)}$ \\
\hline ${E_{m,n,k}}$ & & $\mathit{SOMETIMES(m,n;k)}$\\
\hline
\end{tabular}
\caption{A-ILTL operators}
\label{fig:tableOP}
\end{table}

In the following, we refer to rule-based logic programming languages like DALI,
where A-ILTL formulas occur in the agent program of which they constitute the
check layer. For simplicity, in this context we restrict $\varphi$ to be a conjunction
of literals. Formulas built out of pragmatic A-ILTL operators with such
a restriction on $\varphi$ are called \emph{pragmatic A-ILTL formulas}
(though with some abuse of notation and when clear from the context we will often omit the adjective).
In pragmatic A-ILTL formulas, $\varphi$ must be ground
when the formula is checked. However, similarly to negation-as-failure (where
the negated atom can contain variables, that must however have been instantiated
by literals evaluated previously), we allow variables to occur in an A-ILTL
formula, to be instantiated via a \emph{context} $\chi$.
From the procedural point of view, $\chi$ is required to be evaluated in the first place
so as to make the A-ILTL formula ground.
Notice that, for the evaluation of $\varphi$ and $\chi$, we rely upon the
procedural semantics of the `host' language $\la$. For prolog and DALI,
(extended) resolution procedures \cite{lloyd87} guarantee, with some peculiarities,
correctness and, under some conditions, completeness w.r.t. declarative semantics.
Below, with some abuse of notation we talk about A-ILTL formulas both in a theoretical
and practical sense, in the latter case referring to pragmatic A-ILTL operators with the above restriction
on $\varphi$. Whenever discussing A-ILTL formulas (and, later, A-ILTL expressions and rules)
we will implicitly refer to timed state sequence $\rho_{\ag}$ for given agent,
and by `state(s)' we mean state(s) belonging to this sequence. In practice, this
state sequence will develop along time according to agent's activities,
so during agent operation only the prefix of the state sequence developed up to
the present time can be \as observed'' in order to evaluate A-ILTL formulas.

\begin{definition}
Let $\mathit{OP(m,n;k)\varphi}$ be a pragmatic A-ILTL formula. The corresponding
{\em contextual A-ILTL formula} has the form
$\mathit{OP(M,N;K)\varphi\,::\,\chi}$ where:
\begin{itemize}
\item $M$, $N$ and $K$ can be either variables or constants and $\varphi$ is a conjunction of literals;
\item $\chi$ is called the {\em evaluation context} of the rule, and consists of a
conjunction of literals;
\item each of the $M$, $N$ and $K$ which is a variable and each variable occurring in
$\varphi$ {\em must} occur in an atom (non-negated literal) of $\chi$.
\end{itemize}
\end{definition}

In the following, a contextual A-ILTL formula will implicitly stand for the ground A-ILTL formula
obtained via evaluating the context. We have to establish how to \emph{operationally}
check whether such a formula $\tau$ holds. In fact, during agent operation
one cannot observe the entire state sequence. In all points preceding
the interest interval (as defined in previous section) there is no harm in assuming that $\tau$ holds.
Within the interest interval, $\tau$ can be provisionally assumed to hold if
the inner formula $\varphi$ holds in all points up to now. When the crucial state
(which is the upper bound of the interest interval) is reached, $\tau$ can be definitely established to
hold or not.

\begin{definition}
Given operator $\mathit{OP(m,n)}$ corresponding to A-ILTL operator $\op_{m,n}$
(resp. operator $\mathit{OP(m)}$ corresponding to $\op_{m}$),
an A-ILTL formula $\mathit{OP(m,n)\varphi}$ (resp. $\mathit{OP(m)\varphi}$,
we disregard frequency here)
\emph{operationally holds}
w.r.t. state $s_i$ if, given interval of interest $[v,w]$ of $\op_{m,n}$ (resp. $\op_m$),
one of the following conditions hold:
\begin{itemize}
\item $i < v$;
\item $i \geq v$ and $i\leq w$, i.e., $i$ is in the interest interval, and $\varphi$ holds
(according to the semantics of $\op$)
in all states of sub-interval $[v,i]$;
\item $i \geq w$, i.e., $i$ is the crucial state or $i$ is beyond the crucial state, and $\op\varphi$ holds.
\end{itemize}
\end{definition}

In the next sections, whenever saying that an A-ILTL formula $\tau$ holds we implicitly mean
(unless differently specified or clear from the context) that $\tau$ holds operationally.
For uniformity, the above formulas will be called A-ILTL \emph{rules},
though as we will see below they act as \emph{constraints} that are required
to be fulfilled, otherwise there is an anomaly in the agent's operation.
In what follows we will discuss how to manage such anomalies.

\subsection{A-ILTL Rules with Repair and Improvement}

There can be the case where an A-ILTL expression, checked at a certain
stage of the agent evolution, does not hold (we say that it is \emph{violated}).
What to do upon violation? In static checking, the outcome can
indicate a violation, and the agent program should be modified so as to remove the
anomalous behavior. But, at run-time, no such correction is possible,
and there is in general no user intervention. However, the agent may try to
\emph{repair} itself, by self-modifications to its goals and commitments or even
to its code, by adding/removing (sets of) rules. Even when, on the contrary,
an A-ILTL expression holds, actions may be undertaken as an \emph{improvement}.
Take for instance the example of one who wants to lose some weight by a certain date.
If (s)he fails, then (s)he should undertake a new diet, with less calories.
But if (s)he succeeds before the deadline, then a normocaloric diet should be resumed.

\begin{definition}
An A-ILTL rule with a repair/improvement is a rule the form:
$\mathit{OP(M,N;K)\varphi :: \chi \div \repair \div \improv}$, where:
\begin{itemize}
\item $\mathit{OP(M,N;K)\varphi :: \chi}$ is a contextual A-ILTL rule, called the \emph{monitoring condition};
\item $\repair$ is called the {\em repair action} of the rule, and it consists of an atom $\repair$;
\item $\improv$ (optional) is called the {\em improvement action} of the rule, and it consists of an atom $\repair$.
\end{itemize}
\end{definition}

Whenever the monitoring condition $\mathit{OP(M,N;K)}$ of an A-ILTL
rule is violated, the repair action $\repair$ is attempted. The
repair action is specified via an atom that is `executed'
in the sense that it gives way to an inference process as provided by host language $\la$
(in the prolog terminology, that here we adopt, the atom is a `goal').
If instead the monitoring condition succeeds, in the sense that the specified
interval is expired and the A-ILTL formula holds or, in case of the operator `eventually', if
$\varphi$ holds within given interval, then the improvement action, if specified,
can be `executed'.

The above-mentioned example can be formalized as follows:

\[
\begin{array}{ll}
\mathit{EVENTUALLY(May\!\!-\!\!15\!\!-\!\!2012,June\!\!-\!\!10\!\!-\!\!2012)}\,lose\_five\_kilograms\\
\tbl\div\ \mathit{new\_stricter\_diet(June\!\!-\!\!10\!\!-\!\!2012,June\!\!-\!\!30\!\!-\!\!2012)}\\
\tbl\div\ \mathit{resume\_normal\_diet}
\end{array}
\]

An A-ILTL rule with improvement/repair should of course give way to specified actions
whenever the involved A-ILTL formula can be deemed to hold/not to hold
i.e., in our terminology, as soon as its `critical state' is reached.
Formally:

\begin{definition}
Let $\rho_{\ag}$ be a timed state sequence for agent $\ag$ and let
$\alpha = \tau \div \repair \div \improv$ be an A-ILTL rule with repair/improvement
occurring in $\ag$'s agent program,
where $\tau$ is a contextual A-ILTL formula. $\alpha$ is \emph{fulfilled} in $\rho$
if, given crucial state $\rho_k$ for $\mathit{Op}$, one of the following conditions hold: (i) $\tau$ does not hold, and
$\rho_{k+1} \models \repair$;
(ii) $\tau$ holds, and $\rho_{k+1} \models \improv$.
\end{definition}

\subsection{Evolutionary A-ILTL Expressions}
\label{beh-rul}

It can be useful in many applications to define properties
to be checked upon arrival of partially known sequences of events.
In general in fact,
it is not possible to fully establish in advance which events will arrive
and in which order. Moreover, restricting the agent \as perception'' only
to known events or to an expected order heavily limits the ability of
the agent to improve its behavior in time, e.g. via forms of learning.
This is our motivation for introducing a new kind of A-ILTL rules,
that we call Evolutionary A-ILTL Expressions (first introduced in a preliminary form in \cite{rcra2010,cilc2012}).

These expressions are based upon
specifying: (i) a sequence of past events that may have happened;
(ii) an
A-ILTL formula defining a property that should hold; (iii) a sequence of events
that might happen in the future, without affecting the property;
(iv) a sequence of events
that are supposed \emph{not} to happen in the future, otherwise the property
will not hold any longer;
(v) optionally, \as repair'' actions to be undertaken if the property
is violated.

To be able to indicate in a flexible way
sequences of events of any (unlimited) length
we admit a syntax inspired to regular expressions \cite{Hopcroft:2006}.

\begin{definition}
If $E$ is an event,
$E^{*}$ will indicate zero or more occurrences of $E$, and $E^{+}$ one or more occurrences.
Given events $E_1$ and $E_2$, by $E_1,E2$ we mean that they may
occur in any order; by $E_1 \bullet\bullet E_2$ we mean that $E_1$ must occur before $E_2$ (with possibly
a sequence of unspecified events in between);
by $E_1 \bullet E_2$ we mean that $E_1$ must occur immediately before $E_2$ (i.e., the
two events must be consecutive). Wild-card $X$, standing for unspecified event, can be used
\footnote{Notice that, for an agent, an event \as occurs'' when the agent perceives it.
This is only partially related to when events actually happen in the environment
where the agent is situated. In fact, the order
of perceptions can be influenced by many factors. However, either events are somehow
time-stamped externally (by a reliable third-party) whenever they happen, so as to certify the exact time of their origin
(as sometimes it may be the case), or an agent
must rely on its own subjective experience.}.
Given set of events ${{\cal{E}}v} = \{E_1,\ldots,E_k\}$, $k\geq 0$,
let \emph{an event sequence for (or corresponding to)}
${{\cal{E}}v}$, indicated with ${{\cal{S}}^{{\cal{E}}v}}$, be a sequence defined in the above way on events in ${{\cal{E}}v}$.
Event list $Z_1,\ldots,Z_r$, $r > 0$, \emph{satisfies} ${{\cal{S}}^{{\cal{E}}v}}$ if all
the $Z_i$s occur in ${{\cal{S}}^{{\cal{E}}v}}$ following the specified order.
\end{definition}
For instance, $E_1^{+}\bullet \bullet E_2, E_3 \bullet X \bullet E_4$
means that, after a certain (non-zero) number of occurrences of $E_1$ and, possibly,
of some unknown event, $E_2$ and $E_3$ can occur in any order. They are followed
by one unknown event $X$ and, immediately afterwards, by $E_4$. List $E_1,E_3$ satisfies the
above sequence, as both events occur in it in given order, while list $E_3,E_1$ does not,
as the order is not correct.

\begin{definition}[Evolutionary LTL Expressions]
\label{evexps}
Let ${{\cal{E}}vp} = \{E_{P_1},\ldots,E_{P_l}\}$, $l>1$, be a set of past events,
and ${\cal{F}} = \{F_{1},\ldots,F_{m}\}$, ${\cal{J}} = \{J_{1},\ldots,J_{r}\}$, $m,r \geq 0$,
be sets of events. Let $\calsevp$, $\calsf$ and $\calsj$ be corresponding event sequences.
Let $\tau$ be a contextual A-ILTL formula $\op\,\varphi\,::\,\chi$.
An \emph{Evolutionary LTL Expression} $\varpi$ is of the form $\calsevp\,:\,\tau\,:::\,\calsf\,::::\,\calsj$
where:
\begin{itemize}
\item
$\calsevp$
denote the sequence of relevant events which are supposed to have happened, and in which order,
for the rule to be checked; i.e.,
these events act as preconditions: whenever one or more of them happen in given order, $\tau$ will be checked;
\item
$\calsf$ denote the events that are expected to happen in the future
without affecting $\tau$;
\item
$\calsj$ denote the events that are expected \emph{not} to happen in the future;
i.e., whenever any of them should happen, $\varpi$ is not required to hold any longer, i.e., these can be called are \as breaking events''.
\end{itemize}
The state until which $\varpi$ is required
to hold is the critical state of the operator $\op$ occurring in $\tau$ provided that
if one of the $J_{i}$'s happens at intermediate state $s_{w}$,
then $\varpi$ is not required
to hold after $s_{w}$.
\end{definition}
Notice that both the $F_{i}$'s and the $J_{i}$'s are optional, and that
we do not require the $E_{P_i}$'s, the $F_i$'s  and the $J_i$'s to be ground terms:
variables occurring in them indicate values in which we are not interested.

An Evolutionary LTL Expression can be evaluated w.r.t. a state $s_i$, which contains
(in the component $\varepsilon^{\ag}_i$)
the history $H_i$ of the agent, i.e., the list of past events:
in fact, within an agent, an event has happened if it occurs as a past event in $H_i$. The expression holds,
also in presence of expected or breaking events, if the inner contextual A-ILTL formula $\tau$ holds,
or if a breaking event has occurred (as in this case $\tau$ is no longer required to hold).
Notice that $H_i$ satisfies each of the event sequences in the definition of an
A-ILTL Expression ${\varpi}$ provided that $H_i$ includes zero or more elements of the
sequence, the specified order.
Formally:

\begin{definition}
An Evolutionary A-ILTL Expression ${\varpi}$, of the form specified in Definition\rif{evexps},
\emph{holds} in state $s_i$ whenever (i) $H_i$ satisfies $\calsevp$ and $\calsf$, but not $\calsj$, and $\tau$ holds or
(ii) $H_i$ satisfies $\calsj$.
\end{definition}

\begin{definition}
An Evolutionary A-ILTL Expression ${\varpi}$, of the form specified in Definition\rif{evexps},
is \emph{violated} in state $s_i$ whenever
$H_i$ satisfies $\calsevp$ and $\calsf$, but not $\calsj$,
and $\tau$ does not hold.
\end{definition}

\begin{definition}
An Evolutionary A-ILTL expression ${\varpi}$, of the form specified in Definition\rif{evexps},
is \emph{broken} in state $s_i$ whenever
$H_i$ satisfies $\calsevp$, $\calsf$ and $\calsj$,
and $\tau$ does not hold.
\end{definition}

Operationally, an Evolutionary A-ILTL Expression can be finally deemed to hold if
either the critical state has been reached and $\tau$ holds, or an unwanted event (one of the $J_i$s) has occurred.
Instead, an expression can be deemed \emph{not} to hold (or, as we say,
to be \emph{violated} as far as it expresses a wished-for property) whenever
$\tau$ is false at some point without breaking events.

The following is an example of Evolutionary A-ILTL Expression stating that, after a car has been submitted
to a checkup, it is assumed to work properly for (at least) six months, even in case
of (repeated) long trips, unless an accident occurs.

\[
\begin{array}{ll}
\mathit{checkup_P}(Car)\!:\! T\,:\ \mathit{ALWAYS}(T,T+6_{months})\ \,work\_ok(Car)\\
\tbl:::\mathit{long\_trip}^{+}(Car)\\
\tbl::::\mathit{accident}(Car)
\end{array}
\]

As said before, whenever an unwanted event (one of the $J_i$s) should happen,
$\varpi$ is not required to hold any longer (though it might).
The proposition below formally allows for dynamic run-time checking of Evolutionary A-ILTL Expressions.
In fact, it states that, if a given expression holds in a certain state and is supposed
to keep holding after some expected events have happened, then checking this expression
amounts to checking the modified expression where: (i) the occurred events are removed from event sequences, and (ii) subsequent
events are still expected.

\begin{proposition}
Given Evolutionary A-ILTL Expression of the form specified in Definition\rif{evexps},
assume that $\varpi$ holds at state $s_n$ and that it still holds after the occurrence of event
$E \in {\cal{E}}vp$ and (possibly) of event $F \in {\cal{F}}$
at state $s_v$ ($v \geq n$),
and that none of the events in ${\cal{J}}$ has happened. Let $\calsevpp$ and $\calsff$ be modified event sequences
obtained by respectively canceling $E$ and $F$ from $\calsevp$ and $\calsff$ whenever they occur. Given
${\varpi}_{1}$ $=$ $\calsevpp\,:\,\tau\,:::\,\calsff\,::::\,\calsj$
we have that for every state $s_w$ with ($w\geq v$) $\varpi$ holds iff ${\varpi}_{1}$ holds.
\end{proposition}

Whenever an Evolutionary A-ILTL expression is either violated or broken, a repair can be attempted with the aim
of recovering the agent's state.
\begin{definition}
An evolutionary LTL expression with repair $\varpi^{r}$ is of the form:
\[{\varpi} | \repair_1 || \repair_2\]
where ${\varpi}$ is an Evolutionary LTL Expression adopted in language $\la$, and $\repair_1, \repair_2$ are atoms of $\la$.
$\repair_1$ will be executed
(according to $\la$'s procedural semantics) whenever $\varpi$ is violated, and $\repair_2$ will be executed whenever $\varpi$ is broken.
\end{definition}

\section{Related Work}
\label{comparison}

We may easily notice the similarity between Evolutionary A-ILTL Expressions and event-calculus formulations.
The Event Calculus has been proposed by Kowalski and Sergot
\cite{event_calculus86} as a system for reasoning about time and actions in the
framework of Logic Programming.
The essential idea is to have terms, called {\em fluents,}
which are names of time-dependent relations. Kowalski and Sergot write
$holds(r(x,y),t)$ which is understood as ``fluent $r(x,y)$ is true
at time t''.
Take for instance the default inertia law, stating when fluent $f$ holds,
formulated in the event calculus as follows:

\medskip
$\begin{array}{ll}
holds(f,t) \ \leftarrow & happens(e),\ initiates(e,f),\ date(e,t_s),  \\
                       & t_s < t,\ \no clipped(t_s,f,t)
\end{array} $

\meno
The analogy consists in the fact that,
in the sample A-ILTL expression of previous section, past event $\mathit{checkup_P}(Car):t_1$ initiates a fluent which is actually
an interval A-ILTL expression, namely $G_{t_1,t1+6months}work\_ok(Car)$,
which would be \as clipped'' by $\mathit{accident}(Car)$, where a fluent
which is clipped does not hold any longer. The Evolutionary A-ILTL Expression contains an element
which constitutes an addition w.r.t. the event calculus formulation: in fact,
$\mathit{long\_trip}^{+}(Car)$ represents a sequence of events that is expected, but by which the fluent should \emph{not} be clipped
if everything works as expected. Moreover, in Evolutionary A-ILTL Expressions one can specify a fluent
to initiate and keep holding or terminate according not just to single events, but to complex event sequences of unlimited length.

Static verification of agent programs and systems
(i.e., verification performed prior to agent activation) can be accomplished through model-checking techniques \cite{Clarke:mc}, abstract interpretation
\cite{abstr-int} (not commented here) or theorem proving.

About theorem proving, in \cite{BoerHHM07} for instance, concerning the agent-oriented language GOAL,
a temporal logic is defined to prove properties of GOAL agents. In general, given a logical specification
of an agent and its semantics, properties of that agent can be proved as theorems.

Model-checking is a method for algorithmically checking whether a program
(intended as the \as model'' of a system) satisfies a specification, usually expressed
by means of some kind of temporal logic. In mathematical terms, the method tries to decide if model $M$
(expressed in some formal language), with initial state $s$,
models a property $p$. Otherwise, a counterexample is usually generated. This ia done by exploring all possible state
that given system can possibly reach.
Model-checking techniques \cite{Clarke:mc} have been originally adopted for testing hardware devices, their application to software systems and protocols is constantly growing \cite{spin,mchMAS}, and there have been a number of attempts to overcome some known limitations of this approach.

The application of such techniques to the verification of agents is still limited by two fundamental problems. The first problem arises from the marked differences between the languages used for the definition of agents and those needed by verifiers (usually ad-hoc, tool-specific languages).
Indeed, to apply static verification, currently an agent has to be remodeled in another language: this task is usually performed manually, thus it requires an advanced expertise and gives no guarantee on the correctness and coherence of the new model. In many cases (e.g., \cite{mchMAS,walton}) current research in this field is still focused on the problem of defining a suitable language that can be used to easily and/or automatically reformulate an agent in order to verify it through general model-checking algorithms. For example, \cite{visser2} describes a technique to model-check agents defined by means of a subset of the AgentSpeak language, which can be automatically translated into PROMELA and Java and then verified by the model-checkers SPIN \cite{spin} and Java PathFinder \cite{visser1}, respectively, against a set of constraints which, in turn, are translated into LTL from a source language which is a simplified version of the BDI logic. \cite{Lom2} describes an approach that exploits bounded symbolic model-checking, in particular the tool MCMAS, to check agents and MAS (Multi-Agent Systems) against formulas expressed in the CTLK temporal logic.

The second obstacle is represented by the dynamic nature of agents, which are able to learn and self-modify over their life cycle, and by the extreme variability of the environment in which agents move. These aspects make it difficult to model agents via finite-state languages, which are typical of many model-checkers, and dramatically increase the resources (time, space) required for their verification (state explosion). This can be seen as a motivation for our approach, which defers at least part of the verification activity (namely, the part more dependent upon agent evolution) to run-time.

The literature reports fully-implemented promising verification frameworks (e.g., \cite{visser2,Lom2,sciff-iclp2008}),
of which SCIFF \cite{sciff-iclp2008,sciff-acm2008,sciff2011} is not based upon model-checking.
SCIFF is an abductive proof procedure inspired by the IFF proof procedure \cite{IFF1997} by
Fung and Kowalski.
Unlike original IFF, SCIFF focusses on the externally observable agent behavior.
so as to focalize on the interaction. Agents could be computational entities, reactive systems,
peers in a distributed computer system, even human actors. This kind of detachment
from specific features of an observed system
is called \as social
approach'' to agent interaction.
Given a narrative of such an observed behavior (called a \as history'') the purpose
of the SCIFF framework is (i) to define declaratively whether such a history is
\as admissible'', i.e., compliant to a specification, and (ii) to provide a computational
proof-procedure to decide operationally about its compliance.

In the MAS domain, the SCIFF language has been used
to define agent interaction protocols and, more generally, to describe
the generation of expectations in the form of events, or
\as social goals'', that express the social aim or outcome
of some agent interaction. SCIFF allows one to
model dynamically upcoming events, and specify positive and negative expectations, and the concepts of
fulfilment and violation of expectations.
SCIFF has a declarative semantics given in terms of Abductive Logic Programming,
and is fully implemented. The implementation enjoys important properties, namely
termination, soundness, and completeness w.r.t. the declarative semantics.

Reactive Event Calculus (REC) stems from SCIFF \cite{RECSCIFF2010,REC2011,REC2012}
and exploits the idea that, every time a new event (or set of events)
is delivered to an agent, it must react by extending the narrative and by consequently
extending and revising previously computed results.
REC axiomatization can be based on Abductive Logic Programming
(ALP), or alternatively on a lightweight form of
Cached Event Calculus (CEC) \cite{CEC1996}, that exploits assert and retract
predicates to cache and revise the maximal validity intervals of fluents. This latter semantics is
suitable to deal with application domains where events are processed following
the order in which they have been generated, like in business processes and (web) services.

Another example of social approach is the one based on commitments, firstly introduced by
Singh in \cite{commitments1997}. Commitments result from communicative
actions, and capture the mutual obligations established between
the interacting agents during the execution. \cite{commitments2004} represents
commitments as properties in the event
calculus, and develop a scheme where to model the creation and manipulation
of commitments as a result of performing actions.

All the above-mentioned approaches have relationships with the one presented in this paper.
For instance, positive and negative expectations in SCIFF are similar to the $F$s and $J$s in
Evolutionary A-ILTL Expressions.
However, our approach has its specific original features.
The main one is that it is aimed at a single agent dynamically verifying itself under various respects,
and not to the verification of interactions. Our focus is not on observable behavior
to be confronted with expectations: rather, A-ILTL rules are aimed at expressing inherent agent properties.
We drew inspiration from Cohen and Levesque work on rational agency (see, e.g., \cite{Levesque1988,Cohen1990}).
Let us consider one of their examples from \cite{Cohen1990}, namely \as I always want more money than I have'':
the following is a variation expressed as an Evolutionary A-ILTL Expression, where one
always wants within one month ten percent more money than (s)he has at present. $\mathit{have\_money}_P(S):T$
is a past event that represents the last agent's reminiscence about how much money
(s)he has (in fact, when time-stamp of a past event is a variable, the last version
is obtained). The expression states that by time $T1$, which is $T$ plus one month,
(s)he intends to own a sum $S1$ greater by ten percent than $S$.

\[
\begin{array}{ll}
\mathit{have\_money_P(S)}\!:\! T\,:\\
\tbs \mathit{EVENTUALLY(T1)}\ \mathit{have\_money(S1)}\,::\,S1 = S + 10\%S, T1 = T + 1_{month}
\end{array}
\]

In other approaches, for instance SCIFF or Reactive Event Calculus, one might to some extent specify properties
similar to A-ILTL ones: this however would be easily done
only within agents defined in the related formalisms (respectively, abductive logic programming
and event-calculus), and mechanisms for dynamically checking properties
remain to be defined: in fact, these approaches can be adapted to dynamic checking
when performed by a third party, where they have not been devised for self-checking.
Moreover, the problem of how to apply these methodologies within other
agent-oriented languages and frameworks has still not been considered.
Our approach introduces a flexibility with respect to a pre-defined narrative,
and the concept of repair and improvement is novel.

The approach of \cite{PerlisJLC2005} introduces an \emph{active logic} based on time,
where they explicitly have inference rules such as:

$
\begin{array}{ll}
i: & \mathit{now(i)}\\
& \line(1,0){50}\\
i+1: & \mathit{now(i+1)}
\end{array}
$

\noindent
to denote that the concept of \as present time'' evolves as time passes.
They also have rules for `timed' modus ponens (where the conclusion becomes
true at next instant) and a frame axiom. Our A-ILTL logic has a primitive
`current state' operator that is represented as $\mathit{NOW(t)}$ in the practical syntax,
and can occur in A-ILTL rules and expressions, where it is re-evaluated with
the whole rule/expression at given frequency.

In addition to `base level' systems providing fast reaction,
in \cite{PerlisJLC2005} a component supporting
deliberation and re-consideration,
capable of symbolic reasoning
and meta-reasoning for self-monitoring and self-correction of the overall
system, is advocated. They argue that a flexible, non-brittle system can
be achieved by adding such a layer, where this oversight module
executes a \as Meta-Cognitive loop'' (MCL). Active logic is in their view
a suitable tool for designing MCL, that however should be kept simple and
fast. In a way, our A-ILTL rules and expressions may be seen as composing a sort
of MCL, where A-ILTL can be seen as the underlying active logic.
Then, our approach (especially when applied in the DALI context) could be seen to some extent as an instance of theirs,
with some differences. On the one hand, at the base level they envisage
not necessarily symbolic reasoning modules, but also other kinds
of non-symbolic or purely reactive systems. This of course is a design choice which
is by no means incompatible with A-ILTL. On the other hand, reasoning in time
is possible in DALI also at the base level. In fact, their example of making a date
and meeting somewhere at a certain time might be fully represented by DALI \as internal events'',
which provide rules to be re-evaluated at a certain frequency (`Is it time to go?')
with a reaction that occurs when rule conditions are fulfilled (`If so, then go!').

\section{A Complete Example}
\label{example}

In this section we propose and discuss a small but complete example of use
of the various features that we have introduced so far for agent's dynamic
self-checking. We follow DALI syntax, that we have partly introduced before
and will further explain here. DALI syntax extends prolog syntax, that we suppose
to be known to the reader (cf. e.g., \cite{SterlingS94,lloyd87}.
The example concerns in particular an agent managing a bank cash
machine.

The machine is supposed to be situated in a room. Customers enter and exit
the room via a door. A sensor on the door delivers to the agent the two (external) events
$\mathit{enter\_customer_E}$ and $\mathit{exit\_customer_E}$ when a customer respectively enters or exits
the room. Each external event is managed by a \emph{reactive rule}, where the
traditional prolog connective $\IF$ is replaced by new connective $\THEN$. This new connective
indicates that whenever the event in the head is received (we can also say \as perceived'')
by the agent, then the body is executed.
The agent reacts to a customer entering or exiting by switching on and off the light.
Atoms with postfix $A$ indicate in fact actions. The two actions $\mathit{switch\_on\_light_A}$
and $\mathit{switch\_off\_light_A}$ for simplicity are supposed here non to have preconditions
and to be always successful. After reaction, each external event is automatically recorded as a \emph{past event},
time-stamped with the time of perception/reaction. So, e.g., past event $\mathit{enter\_customer_P: T}$
can be found in the agent's knowledge base whenever external event $\mathit{enter\_customer_E}$
has been perceived at time $T$ (by convention, we take time
of perception to coincide with time of reaction). Whenever such an event
is perceived several times, time-stamp $T$
refers to the last perception (though there is a management of versions,
as illustrated in \cite{CostantiniM11}). Whenever a customer enters the room, the agent expects the customer
to exit within a reasonable time, say $5$ minutes. Otherwise, something irregular may have
happened, so it is better to alert a human operator. All the above is formalized via the
following program fragment. The two reactive rules manage external events.
The subsequent A-ILTL rule states that, if a customer does not follow the expected behavior, then (as a repair
for the violation) the action $\mathit{alert\_operator_A}$ is executed.
Precisely, the rule context (after the $::$) specifies that, if a customer entered at time $T$
(as recorded by past event $\mathit{enter\_customer_P : T}$), the limit time is set to $\mathit{T1 = T +5_m}$.
At most by this time, the user must $\mathit{EVENTUALLY}$ have gone out: i.e.,
past event $\mathit{exit\_customer_P:T2}$ must be found in the agent's knowledge base, recording the user's action of exiting
at a certain time in given interval (in fact, in the context it is stated that $\mathit{T2 > T, T2 \leq T1}$).
The frequency at which this rule is checked is explicitly set at $30$ seconds.

\[
\begin{array}{ll}
\mathit{enter\_customer_E}\ \THEN\ \mathit{switch\_on\_light_A}.\\
\mathit{exit\_customer_E}\ \THEN\ \mathit{switch\_off\_light_A}.\\
\mathit{EVENTUALLY(T, T1, 30_s)}\ \mathit{exit\_customer_P:T2},\, ::\\
\tbl\mathit{enter\_customer_P : T, T1 = T +5_m},\\
\tbl\mathit{T2 > T, T2 \leq T1}\\
\tbl\div\ \mathit{alert\_operator_A}.
\end{array}
\]

\smallskip
When in the room, a customer will presumably attempt to withdraw some money from her/his account.
The customer inserting her/his data in the machine will result
in delivering to the agent an external event such as $\mathit{withdraw(Customer,Sum)_E}$,
where $\mathit{Customer}$ will be instantiated to the customer's code, and $\mathit{Sum}$
to the amount of money that (s)he wants. The reactive rule reported below does the following:
(i) finds the account number $\mathit{Account}$ related to the customer (fails if it does not
exist); (ii) checks whether the customer is trustworthy (fails if withdrawal is for some
security reason prevented); (iii) checks whether the amount is available on the account, and is within the
daily and weakly maximum that can be provided; (iv) updates the customer account by subtracting the sum;
(v) performs an action that will actually result in providing the money. Notice that checking
trust is intended to be an action.

\[
\begin{array}{ll}
\mathit{withdraw(Customer,Sum)_E}\ \THEN\\
\tbl\tbl\mathit{find\_account(Customer,Account),}\\
\tbl\tbl\mathit{check\_trust_A(Customer)},\\
\tbl\tbl\mathit{check\_sum(Sum)},\\
\tbl\tbl\mathit{update(Account,Sum)},\\
\tbl\tbl\mathit{give\_money_A(Account,Sum)}
\end{array}
\]

\smallskip
The two actions occurring in the above reactive rule have preconditions. Preconditions to actions
are expressed via rules where the new connective $\WH$ appears. This means that
the action in the head is enabled (via this rule) to be performed only if the body succeeds.
Actually performing an action then implies the agent being connected to its environment
by practical \emph{actuator} devices. The atom in brackets which (optionally) occurs
in such a rule indicates an alternative action to be performed if the body fails.
The first rule states that money can be provided to the customer only if the
corresponding amount is actually present in the cash machine
(the available balance $B$, that for allowing
the withdrawal must be greater than or equal to required sum, is recorded in the agent knowledge base by fact $\mathit{machine\_content(B)}$).
If not, backup action $\mathit{PrintErrorMsgIW_A}$
will print an error message on the screen of the cash machine indicating that withdrawal is at
the moment impossible.
Before enabling actual withdrawal, balance $B$ must be updated
(by $\mathit{update\_machine\_content(B,B1)}$) by subtracting $\mathit{Sum}$.
The second rule takes trust checking seriously: in fact, if the level of
trust associated to a customer is less than a threshold, then the agent alerts a human operator.
In practice, if trust is sufficiently high then action $\mathit{check\_trust_A(Customer)}$ simply succeeds, otherwise
the backup action will be executed.

\[
\begin{array}{ll}
\mathit{give\_money_A(Sum)}\ \WH\\
\tbl\tbl\mathit{machine\_content(B),B \geq Sum},\\
\tbl\tbl\mathit{update\_machine\_content(B,B1), B1 = B-Sum}\\
\tbl\tbl \{\mathit{PrintErrorMsgIW_A}\}.\\
\mathit{check\_trust_A(Customer)}\ \WH\\
\tbl\tbl\mathit{trust(Customer,L), L > Threshold}\\
\tbl\tbl\{Alert\_Operator_A\}.
\end{array}
\]

\smallskip
Finally, the agent is responsible of providing the cash machine with money to be given to customers.
The agent fills the machine with a standard quantity $Q$ of money.
This quantity is recorded in the agent's knowledge base by fact $standard\_quantity(Q)$.
The filling is performed in practice by an action such as $\mathit{fill\_machine_A(Q)}$, executed at time $T$.
Each action, similarly to external events, after execution is recorded in the agent's knowledge base
as a past event: in this case, a past event will be added of the form $\mathit{fill\_machine_P(Q):T}$.
After filling the machine, the agent expects the money to be sufficient for some time,
say $8$ hours.

The following Evolutionary A-ILTL expression states that, after the action
of filling the machine, for eight hours the machine
should not get empty.
Precisely, the time interval considered in the expression is $[T,T1]$ where,
as specified in the evaluation context of the rule, $T1 = T+8_{hours}$. Within this interval,
the machine content (recorded in the agent's knowledge base by fact $\mathit{(machine\_content(B)}$) must be
$\mathit{ALWAYS}$, i.e., in all time instants of the interval, greater than a $\mathit{minimum}$ amount.
This obviously also in case (as stated after the $:::$) of repeated withdrawals
(whatever their number) though,
as seen before, each withdrawal makes the machine content $B$ decrease. In case of violation,
i.e., in case the machine gets empty or almost (content less than a $\mathit{minimum}$ amount),
the agent (as stated after the $|$) fills again the machine with standard quantity $Q$ of money.
This again by means of action $\mathit{fill\_machine_A(Q)}$, that will in turn become a past event with a new
time-stamp corresponding to its completion. The agent however will also reconsider the standard quantity,
as $Q$ has proven to be insufficient,
possibly updating it to a (presumably larger) new amount $Q1$. All this unless (as specified after the $::::$) an exceptional condition,
in this case robbery, occurs. If so, the repair action executed is (as specified after the $||$) to call the police.
Notice that an A-ILTL expression is checked at a certain frequency (in this case the default one).
So, it can be the case that the condition is violated but the violation has not been detected yet.
This case however is managed by rules coping with withdrawal: in particular, if money cannot be provided
because it is insufficient, as seen before an error message will be displayed.

\[
\begin{array}{ll}
\mathit{fill\_machine_P(Q)}\!:\! T\ :\ \mathit{ALWAYS}({T,T1)}\\
\tbl\mathit{(machine\_content(B),B > minimum)\,::\, T1 = T+8_{hours}}\\
\tbl\tbl:::\,\mathit{withdraw(A,S)_A}^{+}\ |\\
\tbl\tbl\tbl\tbs\ \mathit{standard\_quantity(Q)},\mathit{fill\_machine_A(Q)},\\
\tbl\tbl\tbl\tbs\ \mathit{reconsider\_quantity(Q,Q1)}\\
\tbl\tbl::::\mathit{robbery}\,||\,\mathit{call\_police_A}.
\end{array}
\]

\section{Concluding Remarks}
\label{end}
In this paper, we have presented a comprehensive framework for defining
agent properties and run-time self-checking of such properties. Our methodology is
intended as complementary to static verification techniques. To this
aim, we have introduced A-ILTL rules and expressions, that allow one to define a number
of useful properties that the evolution of an agent should fulfil, whatever the
sequence of in-coming events. We have provided a complete semantic framework,
adaptable to several practical settings. The approach is significantly different from
related work, to which it could to some extent be usefully integrated.
The approach has been prototypically implemented and experimented \cite{cilc2012} in the context of the DALI language.

\section*{Acknowledgement}
\label{ack}
My mentor Gaetano Aurelio Lanzarone (`Elio' for his friends) died at the age of 66 in October 2011 after a long illness.
He has been one of the pioneers of prolog
and rule-based automated reasoning in Italy, and initiated me into this fascinating research area.
My affection and gratitude to Elio will last forever. This paper is dedicated to him.
I gratefully acknowledge past joint work and a long-lasting friendship and cooperation on the subjects discussed
in the present paper with Pierangelo Dell'Acqua, Lu\'{\i}s Moniz Pereira, Arianna Tocchio and  Francesca Toni.

%


\end{sloppypar}

\begin{thebibliography}{10}
	
	\bibitem{survey-logag07}
	Fisher, M., Bordini, R.H., Hirsch, B., Torroni, P.:
	\newblock Computational logics and agents: a road map of current technologies
	and future trends.
	\newblock Computational Intelligence Journal \textbf{23}(1) (2007)  61--91
	
	\bibitem{Wooldridge01}
	Wooldridge, M., Dunne, P.E.:
	\newblock The computational complexity of agent verification.
	\newblock In: Intelligent Agents VIII, 8th International Workshop, ATAL 2001
	Seattle, WA, USA, August 1-3, 2001, Revised Papers. Volume 2333 of Lecture
	Notes in Computer Science., Springer (2002)
	
	\bibitem{Clarke:mc}
	Clarke, E.M., Lerda, F.:
	\newblock Model checking: Software and beyond.
	\newblock Journal of Universal Computer Science \textbf{13}(5) (2007)  639--649
	
	\bibitem{abstr-int}
	Cousot, P., Cousot, R.:
	\newblock Abstract interpretation: a unified lattice model for static analysis
	of programs by construction or approximation of fixpoints.
	\newblock In: Conference Record of the Fourth Annual ACM SIGPLAN-SIGACT
	Symposium on Principles of Programming Languages, Los Angeles, California,
	ACM Press, New York, NY (1977)  238--252
	
	\bibitem{Brachman06}
	Brachman, R.J.:
	\newblock ({A}{A}){A}{I} more than the sum of its parts.
	\newblock AI Magazine \textbf{27}(4) (2006)  19--34
	
	\bibitem{PerlisJLC2005}
	Anderson, M.L., Perlis, D.:
	\newblock Logic, self-awareness and self-improvement: the metacognitive loop
	and the problem of brittleness.
	\newblock J. Log. Comput. \textbf{15}(1) (2005)  21--40
	
	\bibitem{AndersonFJOPWW08}
	Anderson, M.L., Fults, S., Josyula, D.P., Oates, T., Perlis, D., Wilson, S.,
	Wright, D.:
	\newblock A self-help guide for autonomous systems.
	\newblock AI Magazine \textbf{29}(2) (2008)  67--73
	
	\bibitem{postdalt06}
	Costantini, S., Tocchio, A.:
	\newblock About declarative semantics of logic-based agent languages.
	\newblock In Baldoni, M., Endriss, U., Omicini, A., Torroni, P., eds.:
	Declarative Agent Languages and Technologies III, Third International
	Workshop, DALT 2005, Selected and Revised Papers. Volume 3904 of LNAI.
	\newblock Springer (2006)  106--123
	
	\bibitem{CostantiniM11}
	Costantini, S.:
	\newblock Defining and maintaining agent's experience in logical agents.
	\newblock In: Informal Proc. of the LPMAS (Logic Programming for Multi-Agent
	Systems) Workshop at ICLP 2011, and CORR Proceedings of LANMR 2011,
	Latin-American Conference on Non-Monotonic Reasoning. (2011)
	
	\bibitem{costandper}
	Costantini, S., Dell'Acqua, P., Pereira, L.M.:
	\newblock A multi-layer framework for evolving and learning agents.
	\newblock In M.~T.~Cox, A.R., ed.: Proceedings of Metareasoning: Thinking about
	thinking workshop at AAAI 2008, Chicago, USA. (2008)
	
	\bibitem{jelia02}
	Costantini, S., Tocchio, A.:
	\newblock A logic programming language for multi-agent systems.
	\newblock In: Logics in Artificial Intelligence, Proc. of the 8th Europ.
	Conf.,JELIA 2002. LNAI 2424, Springer-Verlag, Berlin (2002)
	
	\bibitem{jelia04}
	Costantini, S., Tocchio, A.:
	\newblock The {DALI} logic programming agent-oriented language.
	\newblock In: Logics in Artificial Intelligence, Proc. of the 9th European
	Conference, Jelia 2004. LNAI 3229, Springer-Verlag, Berlin (2004)
	
	\bibitem{dali-references}
	Costantini, S.:
	\newblock The {D}{A}{L}{I} agent-oriented logic programming language:
	References (2012) at URL http://www.di.univaq.it/stefcost/info.htm.
	
	\bibitem{webDALI}
	Costantini, S., D'Alessandro, S., Lanti, D., Tocchio, A.:
	\newblock Dali web site, download of the interpreter (2010)
	http://www.di.univaq.it/stefcost/Sito-Web-DALI/WEB-DALI/index.php, With the
	contribution of many undergraduate and graduate students of Computer Science,
	L'Aquila. For beta-test versions of the interpreter (latest advancements)
	please ask the authors.
	
	\bibitem{LIRA}
	Castaldi, M., Costantini, S., Gentile, S., Tocchio, A.:
	\newblock A logic-based infrastructure for reconfiguring applications.
	\newblock In Leite, J.A., Omicini, A., Sterling, L., Torroni, P., eds.:
	Declarative Agent Languages and Technologies, First International Workshop,
	DALT 2003, Revised Selected and Invited Papers. Volume 2990 of LNAI.,
	Springer (2004) Hot Topics Sub-series.
	
	\bibitem{woa07}
	Costantini, S., Mostarda, L., Tocchio, A., Tsintza, P.:
	\newblock Agents and security in a cultural assets transport scenario.
	\newblock In: Agents and Industry: Technological Applications of Software
	Agents, Proc. of WOA'07
	
	\bibitem{CMTT08}
	Costantini, S., Mostarda, L., Tocchio, A., Tsintza, P.:
	\newblock Dalica agents applied to a cultural heritage scenario.
	\newblock IEEE Intelligent Systems, Special Issue on Ambient Intelligence
	\textbf{23}(8) (2008)
	
	\bibitem{Costantini-PAAMS12}
	Bevar, V., Costantini, S., Tocchio, A., Gasperis, G.D.:
	\newblock A multi-agent system for industrial fault detection and repair.
	\newblock In Demazeau, Y., M{\"u}ller, J.P., Rodr\'{\i}guez, J.M.C., P{\'e}rez,
	J.B., eds.: Advances on Practical Applications of Agents and Multi-Agent
	Systems - Proc. of PAAMS 2012. Volume 155 of Advances in Soft Computing.,
	Springer (2012)  47--55 Related Demo paper ``Demonstrator of a Multi-Agent
	System for Industrial Fault Detection and Repair'', pages 237-240 of same
	volume.
	
	\bibitem{soar:comparison}
	{SOAR-Research-Group}:
	\newblock {SOAR}: A comparison with rule-based systems (2010) URL:
	http://sitemaker.umich.edu/soar/home.
	
	\bibitem{Pnueli77}
	Pnueli, A.:
	\newblock The temporal logic of programs.
	\newblock In: Proc. of FOCS, 18th Annual Symposium on Foundations of Computer
	Science, IEEE (1977)  46--57
	
	\bibitem{tl3}
	Emerson, E.A.:
	\newblock Temporal and modal logic.
	\newblock In van Leeuwen, J., ed.: Handbook of Theoretical Computer Science,
	vol. B.
	\newblock MIT Press (1990)
	
	\bibitem{CDP09}
	Costantini, S., Dell'Acqua, P., Pereira, L.M., Tsintza, P.:
	\newblock Runtime verification of agent properties.
	\newblock In: Proc. of the Int. Conf. on Applications of Declarative
	Programming and Knowledge Management (INAP09). (2009)
	
	\bibitem{Koymans90}
	Koymans, R.:
	\newblock Specifying real-time properties with metric temporal logic.
	\newblock Real-Time Systems \textbf{2}(4) (1990)  255--299
	
	\bibitem{AlurH91}
	Alur, R., Henzinger, T.A.:
	\newblock Logics and models of real time: A survey.
	\newblock In de~Bakker, J.W., Huizing, C., de~Roever, W.P., Rozenberg, G.,
	eds.: Real-Time: Theory in Practice, REX Workshop, Mook, The Netherlands,
	June 3-7, 1991, Proceedings. Volume 600 of Lecture Notes in Computer
	Science., Springer (1992)  74--106
	
	\bibitem{HirshfeldR04}
	Hirshfeld, Y., Rabinovich, A.M.:
	\newblock Logics for real time: Decidability and complexity.
	\newblock Fundam. Inform. \textbf{62}(1) (2004)  1--28
	
	\bibitem{itl2004}
	Goranko, V., Montanari, A., Sciavicco, G.:
	\newblock A road map of interval temporal logics and duration calculi.
	\newblock Journal of Applied Non-Classical Logics \textbf{14}(1-2) (2004)
	9--54
	
	\bibitem{Manna2010}
	Manna, Z., Pnueli, A.:
	\newblock Temporal verification of reactive systems: Response.
	\newblock In: Time for Verification, Essays in Memory of Amir Pnueli. Volume
	6200 of Lecture Notes in Computer Science., Springer (2010)  279--361
	
	\bibitem{barringer89metatem}
	Barringer, H., Fisher, M., Gabbay, D., Gough, G., Owens, R.:
	\newblock Metate{M}: {A} framework for programming in temporal logic.
	\newblock In: Proceedings of REX Workshop on Stepwise Refinement of Distributed
	Systems: Models, Formalisms, Correctness. LNCS 430, Springer-Verlag (1989)
	
	\bibitem{Fisher05}
	Fisher, M.:
	\newblock Metate{M}: The story so far.
	\newblock In Bordini, R.H., Dastani, M., Dix, J., Fallah-Seghrouchni, A.E.,
	eds.: PROMAS. LNCS 3862, Springer (2005)  3--22
	
	\bibitem{metatem1995}
	Barringer, H., Fisher, M., Gabbay, D.M., Gough, G., Owens, R.:
	\newblock Metate{M}: An introduction.
	\newblock Formal Asp. Comput. \textbf{7}(5) (1995)  533--549
	
	\bibitem{GovernatoriR11}
	Governatori, G., Rotolo, A.:
	\newblock Justice delayed is justice denied: Logics for a temporal account of
	reparations and legal compliance.
	\newblock In: CLIMA. Volume 6814 of Lecture Notes in Computer Science.,
	Springer (2011)  364--382
	
	\bibitem{Manna1991}
	Henzinger, T.A., Manna, Z., Pnueli, A.:
	\newblock Timed transition systems.
	\newblock In de~Bakker, J.W., Huizing, C., de~Roever, W.P., Rozenberg, G.,
	eds.: Real-Time: Theory in Practice, REX Workshop, Mook, The Netherlands,
	June 3-7, 1991, Proceedings. Volume 600 of Lecture Notes in Computer
	Science., Springer (1992)  226--251
	
	\bibitem{Gelfond07}
	Gelfond, M.:
	\newblock Answer sets.
	\newblock In: Handbook of Knowledge Representation, Chapter 7.
	\newblock Elsevier (2007)
	
	\bibitem{Manna1984}
	Manna, Z., Pnueli, A.:
	\newblock Adequate proof principles for invariance and liveness properties of
	concurrent programs.
	\newblock Sci. Comput. Program. \textbf{4}(3) (1984)  257--289
	
	\bibitem{jelia02:evolp}
	Alferes, J.J., Brogi, A., Leite, J.A., Pereira, L.M.:
	\newblock Evolving logic programs.
	\newblock In: Logics in Artificial Intelligence, Proc. of the 8th Europ. Conf.,
	JELIA 2002. LNAI 2424, Springer-Verlag, Berlin (2002)  50--61
	
	\bibitem{epia03:evolp}
	J.Alferes, J., Brogi, A., Leite, J.A., Pereira, L.M.:
	\newblock An evolvable rule-based e-mail agent.
	\newblock In: Procs. of the 11th Portuguese Intl.Conf. on Artificial
	Intelligence (EPIA'03). LNAI 2902, Springer-Verlag, Berlin (2003)  394--408
	
	\bibitem{CDP2011}
	Costantini, S., Dell'Acqua, P., Pereira, L.M.:
	\newblock Conditional learning of rules and plans by knowledge exchange in
	logical agents.
	\newblock In: Proc. of Rule{M}{L} 2011 at IJCAI. (2011)
	
	\bibitem{bcdl:jlc00}
	Barklund, J., Dell'Acqua, P., Costantini, S., Lanzarone, G.A.:
	\newblock Reflection principles in computational logic.
	\newblock J. of Logic and Computation \textbf{10}(6) (2000)  743--786
	
	\bibitem{LTL-complexity1985}
	Sistla, A.P., Clarke, E.M.:
	\newblock The complexity of propositional linear temporal logics.
	\newblock J. ACM \textbf{32}(3) (1985)  733--749
	
	\bibitem{kowalgs1979}
	Kowalski, R.A.:
	\newblock Algorithm = logic + control.
	\newblock Commun. ACM \textbf{22}(7) (1979)  424--436
	
	\bibitem{lloyd87}
	Lloyd, J.W.:
	\newblock Foundations of Logic Programming.
	\newblock Springer (1987)
	
	\bibitem{rcra2010}
	Costantini, S., Dell'Acqua, P., Pereira, L.M., Tocchio, A.:
	\newblock Ensuring agent properties under arbitrary sequences of incoming
	events.
	\newblock In: Proc. of 17th {RCRA} Intl. Worksh. on Experimental evaluation of
	algorithms for solving problems with combinatorial explosion. (2010)
	
	\bibitem{cilc2012}
	Costantini, S., Tsintza, P.:
	\newblock Temporal meta-axioms in logical agents.
	\newblock In: Electr. Proc. of CILC 2012, Italian Conference of Computational
	Logic. Number 857 in CEUR Workshop Proceedings Series (2012)
	
	\bibitem{Hopcroft:2006}
	Hopcroft, J.E., Motwani, R., Ullman, J.D.:
	\newblock Introduction to Automata Theory, Languages, and Computation (3rd
	Edition).
	\newblock Addison-Wesley Longman Publishing Co., Inc., Boston, MA, USA (2006)
	
	\bibitem{event_calculus86}
	Kowalski, R., Sergot, M.:
	\newblock A logic-based calculus of events.
	\newblock New Generation Computing \textbf{4} (1986)  67--95
	
	\bibitem{BoerHHM07}
	de~Boer, F.S., Hindriks, K.V., van~der Hoek, W., Meyer, J.J.C.:
	\newblock A verification framework for agent programming with declarative
	goals.
	\newblock J. Applied Logic \textbf{5}(2) (2007)  277--302
	
	\bibitem{spin}
	Holzmann, G.:
	\newblock The model checker spin.
	\newblock IEEE Transactions on Software Engineering (23) (199)  279--295
	
	\bibitem{mchMAS}
	Bourahla, M., Benmohamed, M.:
	\newblock Model checking multi-agent systems.
	\newblock Informatica (Slovenia) \textbf{29}(2) (2005)  189--198
	
	\bibitem{walton}
	Walton, C.:
	\newblock Verifiable agent dialogues.
	\newblock J. Applied Logic \textbf{5}(2) (2007)  197--213
	
	\bibitem{visser2}
	Bordini, R., Fisher, M., Visser, W., Wooldridge, M.:
	\newblock Verifying multi-agent programs by model checking.
	\newblock Autonomous Agents and Multi-Agent Systems \textbf{12}(2) (2006)
	239--256
	
	\bibitem{visser1}
	Visser, W., Havelund, K., Brat, G., Park, S., Lerda, F.:
	\newblock Model checking programs.
	\newblock Autom. Softw. Eng.
	
	\bibitem{Lom2}
	Jones, A., Lomuscio, A.:
	\newblock Distributed bdd-based bmc for the verification of multi-agent
	systems.
	\newblock In: Proc. of the 9th Int. Conf. on Autonomous Agents and Multiagent
	Systems (AAMAS 2010). (2010)
	
	\bibitem{sciff-iclp2008}
	Montali, M., Alberti, M., Chesani, F., Gavanelli, M., Lamma, E., Mello, P.,
	Torroni, P.:
	\newblock Verification from declarative specifications using logic programming.
	\newblock In: 24th Int. Conf. on Logic Programming (ICLP'08). Volume 5366 of
	Lecture Notes in Computer Science., Springer (2008)  440--454
	
	\bibitem{sciff-acm2008}
	Alberti, M., Chesani, F., Gavanelli, M., Lamma, E., Mello, P., Torroni, P.:
	\newblock Verifiable agent interaction in abductive logic programming: The
	sciff framework.
	\newblock ACM Trans. Comput. Log. \textbf{9}(4) (2008)
	
	\bibitem{sciff2011}
	Montali, M., Chesani, F., Mello, P., Torroni, P.:
	\newblock Modeling and verifying business processes and choreographies through
	the abductive proof procedure sciff and its extensions.
	\newblock Intelligenza Artificiale, Intl. J. of the Italian Association AI*IA
	\textbf{5}(1) (2011)
	
	\bibitem{IFF1997}
	Fung, T.H., Kowalski, R.A.:
	\newblock The {I}{F}{F} proof procedure for abductive logic programming.
	\newblock J. Log. Program. \textbf{33}(2) (1997)  151--165
	
	\bibitem{RECSCIFF2010}
	Chesani, F., Mello, P., Montali, M., Torroni, P.:
	\newblock A logic-based, reactive calculus of events.
	\newblock Fundam. Inform. \textbf{105}(1-2) (2010)  135--161
	
	\bibitem{REC2011}
	Torroni, P., Chesani, F., Mello, P., Montali, M.:
	\newblock A retrospective on the reactive event calculus and commitment
	modeling language.
	\newblock In Sakama, C., Sardi{\~n}a, S., Vasconcelos, W., Winikoff, M., eds.:
	Declarative Agent Languages and Technologies IX - 9th International Workshop,
	DALT 2011, Revised Selected and Invited Papers. Volume 7169 of Lecture Notes
	in Computer Science., Springer (2012)  120--127
	
	\bibitem{REC2012}
	Bragaglia, S., Chesani, F., Mello, P., Montali, M., Torroni, P.:
	\newblock Reactive event calculus for monitoring global computing applications.
	\newblock In Artikis, A., Craven, R., Cicekli, N.K., Sadighi, B., Stathis, K.,
	eds.: Logic Programs, Norms and Action - Essays in Honor of Marek J. Sergot
	on the Occasion of His 60th Birthday. Volume 7360 of Lecture Notes in
	Computer Science., Springer (2012)  123--146
	
	\bibitem{CEC1996}
	Chittaro, L., Montanari, A.:
	\newblock Efficient temporal reasoning in the cached event calculus.
	\newblock Computational Intelligence \textbf{12} (1996)  359--382
	
	\bibitem{commitments1997}
	Singh, M.P.:
	\newblock Commitments in the architecture of a limited, rational agent.
	\newblock In Cavedon, L., Rao, A.S., Wobcke, W., eds.: Intelligent Agent
	Systems, Theoretical and Practical Issues, Based on a Workshop Held at
	PRICAI'96. Volume 1209 of Lecture Notes in Computer Science., Springer (1997)
	72--87
	
	\bibitem{commitments2004}
	Yolum, P., Singh, M.P.:
	\newblock Reasoning about commitments in the event calculus: An approach for
	specifying and executing protocols.
	\newblock Ann. Math. Artif. Intell. \textbf{42}(1-3) (2004)  227--253
	
	\bibitem{Levesque1988}
	Levesque, H.J.:
	\newblock Comments on "knowledge, representation, and rational
	self-government".
	\newblock In Vardi, M.Y., ed.: TARK, Morgan Kaufmann (1988)  361--362
	
	\bibitem{Cohen1990}
	Cohen, P.R., Levesque, H.J.:
	\newblock Intention is choice with commitment.
	\newblock Artif. Intell. \textbf{42}(2-3) (1990)  213--261
	
	\bibitem{SterlingS94}
	Sterling, L., Shapiro, E.Y.:
	\newblock The Art of Prolog - Advanced Programming Techniques, 2nd Ed.
	\newblock MIT Press (1994)
	
\end{thebibliography}
\end{document}